\documentclass[]{fairmeta}
\usepackage{wrapfig}
\usepackage{tabularx}
\usepackage{textcomp}
\usepackage{stfloats}
\usepackage{url}
\usepackage{verbatim}
\usepackage{titlesec}
\usepackage{adjustbox}
\usepackage{multirow}
\usepackage{pifont}
\usepackage{comment}
\usepackage{amsmath,amssymb} % define this before the line numbering.
\usepackage{color}
\usepackage{hyperref}
\usepackage{graphicx}    % 用于插入图片
\usepackage{longtable}
\usepackage{booktabs} % 表格更好的线条
\usepackage{subcaption} % 支持 subtable 和 subfigure

\RequirePackage{xspace}
\makeatletter
\DeclareRobustCommand\onedot{\futurelet\@let@token\@onedot}
\def\@onedot{\ifx\@let@token.\else.\null\fi\xspace}

\makeatother

\definecolor{adptorange}{RGB}{248, 205, 172}
\definecolor{cmpblue}{RGB}{189, 215, 238}
\definecolor{cmpblue}{RGB}{189, 215, 238}

\definecolor{our_red}{RGB}{232,157,160}
\definecolor{our_blue}{RGB}{136,206,230}
\definecolor{our_orange}{RGB}{246,200,168}
\definecolor{our_green}{RGB}{178,211,164}

\definecolor{attn_code0}{RGB}{247,215,200}
\definecolor{attn_code1}{RGB}{238,169,139}
\definecolor{mlp_code0}{RGB}{204,201,221}
\definecolor{mlp_code1}{RGB}{102,95,153}

\definecolor{token_blue}{RGB}{84, 120, 140}
\usepackage{pifont}       % \ding{xx}
\usepackage{bbding}       % \Checkmark,\XSolid,... (需要和pifont宏包共同使用)
\usepackage{fontawesome}
\usepackage{xspace}

\usepackage{float}

\usepackage{array}

\newlength\savewidth

\newcolumntype{x}[1]{>{\centering\arraybackslash}p{#1pt}}
\newcolumntype{y}[1]{>{\raggedright\arraybackslash}p{#1pt}}
\newcolumntype{z}[1]{>{\raggedleft\arraybackslash}p{#1pt}}

\renewcommand{\paragraph}[1]{\vspace{1mm}\noindent\textbf{#1}}
\usepackage{colortbl}
\usepackage{xcolor}
\renewcommand{\paragraph}[1]{\vspace{1.25mm}\noindent\textbf{#1}}

\usepackage{algorithm}
\usepackage{listings}

\definecolor{codeblue}{rgb}{0.25, 0.5, 0.5}
\definecolor{codekw}{rgb}{0.35, 0.35, 0.75}
\lstdefinestyle{Pytorch}{
    language = Python,
    backgroundcolor = \color{white},
    basicstyle = \fontsize{9pt}{8pt}\selectfont\ttfamily\bfseries,
    columns = fullflexible,
    aboveskip=1pt,
    belowskip=1pt,
    breaklines = true,
    captionpos = b,
    commentstyle = \color{codeblue},
    keywordstyle = \color{codekw},
}

\definecolor{green}{HTML}{009000}
\definecolor{red}{HTML}{ea4335}

\title{TumorChain: Interleaved Multimodal Chain-of-Thought Reasoning for Traceable Clinical Tumor Analysis}

\author{Sijing Li$^{1,2*}$, Zhongwei Qiu$^{2,3,1*}$, Jiang Liu$^1$, Wenqiao Zhang$^{1\dagger}$, Tianwei Lin$^{1,2}$,
\textbf{Yihan Xie}$^1$, \textbf{Jianxiang An}$^1$, \\ \textbf{Boxiang Yun}$^2$, \textbf{Chenglin Yang}$^1$, \textbf{Jun Xiao}$^1$,
\textbf{Guangyu Guo}$^{2,1}$, \textbf{Jiawen Yao}$^{2}$, \textbf{Wei Liu}$^2$, \textbf{Yuan gao}$^{2}$, \textbf{Ke Yan}$^{2}$, \textbf{Weiwei Cao}$^{2}$,
\textbf{Zhilin Zheng}$^{2}$, \textbf{Tony C. W. MOK}$^{2}$, \textbf{Kai Cao}$^4$, \textbf{Yu Shi}$^{5}$, \textbf{Jiuyu Zhang}$^5$, \textbf{Jian Zhou}$^{6}$,\\
\textbf{Beng Chin Ooi}$^1$, \textbf{Yingda Xia}$^{2\dagger}$, \textbf{Ling Zhang}$^{2}$\\

\affiliation[1]{Zhejiang University}
\affiliation[2]{DAMO Academy, Alibaba Group}
\affiliation[3]{Hupan Lab\\}
\affiliation[4]{Shanghai Institution of Pancreatic Disease}
\affiliation[5]{Shengjing Hospital of China Medical University\\}
\affiliation[6]{Sun Yat-sen University Cancer Center}

\texttt{\{suzylee, wenqiaozhang\}@zju.edu.cn},\\ \texttt{\{qiuzhongwei.qzw, yingda.xia\}@alibaba-inc.com}\\
}
\abstract{
Accurate tumor analysis is central to clinical radiology and precision oncology, where early detection, reliable lesion characterization, and pathology-level risk assessment directly guide diagnosis, staging, and treatment planning. Chain-of-Thought (CoT) reasoning is particularly critical in this setting, as it enables stepwise interpretation from imaging findings to clinical impressions and pathology-level conclusions, ensuring traceability and reducing diagnostic errors. Here, we target the clinical tumor analysis task and build a large-scale benchmark that operationalizes a multimodal reasoning pipeline, spanning findings, impressions, and pathology predictions. 
We curate \texttt{TumorCoT}, a large-scale dataset of \textbf{1.5M CoT-labeled VQA instructions} paired with 3D CT scans, with step-aligned rationales and cross-modal alignments along the “findings → impression → pathology” trajectory, enabling standardized evaluation of both final accuracy and reasoning consistency.
We further propose \textbf{TumorChain}, a multimodal interleaved reasoning framework that tightly couples 3D imaging encoders, clinical text understanding, and organ-level vision-language alignment. 
Through cross-modal alignment and iterative interleaved causal reasoning, TumorChain grounds visual evidence, aggregates conclusions, and issues pathology predictions after multiple rounds of self-refinement, improving traceability and reducing hallucination risk. 
TumorChain demonstrates consistent gains over strong unimodal and pipeline baselines in lesion detection, impression quality, and pathology classification, and successfully generalizes to the public DeepTumorVQA benchmark. Ablations validate the key contributions of interleaved reasoning and clinical CoT. Clinically, these advances lay the groundwork for reliable, interpretable tumor assessment to support real-world decision-making. To advance safe, explainable, and reproducible multimodal reasoning for high-stakes tumor analysis, detailed information about our project can be found on our project homepage at \href{https://github.com/ZJU4HealthCare/TumorChain}{https://github.com/ZJU4HealthCare/TumorChain}.
}

\date{\today}

\begin{document}
\thispagestyle{firstheader}
\maketitle
\pagestyle{empty}

\section{introduction}

\begin{figure*}[t]
  \centering
  \includegraphics[width=1\linewidth,clip=false, trim=0 15 5 0]{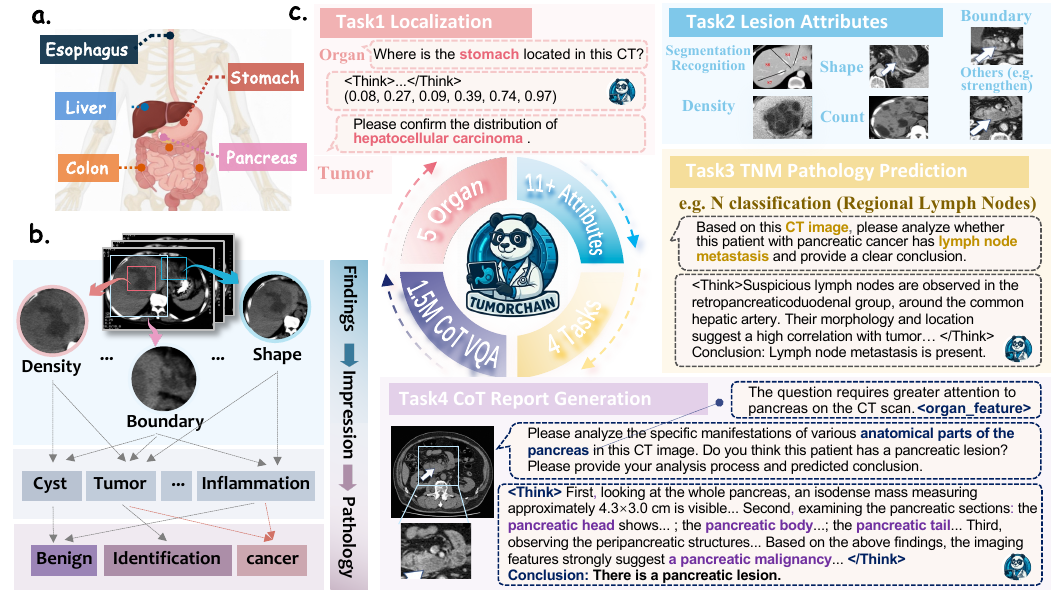}
    \vspace{3mm}
  \caption{(a) \textbf{TumorChain} focuses on tumor analysis of the five major organs in the digestive system. (b) We establish a corresponding medical reasoning logic chain from radiology findings to radiology impressions to pathological conclusions. (c) The system tackles four diverse tasks across 1.5 million multimodal VQA instructions, enabling advanced clinical reasoning.}
    \vspace{-6mm}
  \label{fig:intro}
\end{figure*}

\renewcommand{\thefootnote}{\fnsymbol{footnote}} % 让脚注编号变成 *, †, ‡ ...
\footnotetext[1]{Equal contributions. The work was done during Sijing’s internship at DAMO Academy.}           % [1] 对应第一个符号：*
\footnotetext[2]{Corresponding author.}           % [1] 对应第一个符号：*
\renewcommand{\thefootnote}{\arabic{footnote}}   % 需要的话再改回数字
% Foundation models have progressed at an unprecedented rate, transitioning from large language models (LLMs)~\citep{brown2020language,openai2023gpt4} to multimodal architectures~\citep{radford2021learning} that demonstrate broad applicability across diverse domains. Within healthcare, these advances have equipped models with the capacity to interpret complex clinical data, integrate heterogeneous modalities, and provide decision support, creating new opportunities to enhance screening~\citep{hu2025ai}, diagnosis~\citep{shui2025large}, treatment~\citep{lu2024multimodal}, and patient care~\citep{lin2025healthgpt,qiu2024llm}.

% 尽管基础模型在临床领域取得了显著成功~\citep{zhang2023huatuogpt,wang2025baichuan}，但其在跨模态临床理解与推理方面仍面临诸多挑战，尤其是在基于影像组学的肿瘤发现与风险评估上。例如，肿瘤科医生需要在影像中识别器官肿瘤，区分良恶性病变，并结合病理信息及其他多维患者特征，评估TNM分期~\footnote{TNM系统用于描述患者体内癌症的范围和扩散情况。T代表肿瘤的大小及其向邻近组织的蔓延；N代表癌症向附近淋巴结的扩散；M代表癌症向身体其他部位的转移（转移灶）。}~\citep{amin2017ajcc}及癌症风险，以制定治疗方案。
Although large vision-language models (LVLMs) have achieved considerable success in general healthcare domains~\citep{yang2024advancing,lin2025healthgpt,xu2025lingshu,lai2025med}, their application to multi-modal clinical understanding and reasoning remains challenging, especially in cancer discovery and risk assessment~\citep{qiu2025bridging} based on medical images. 
For example, oncologists need to identify suspected tumors on radiological images, distinguish benign from malignant lesions, and assess cancer risk and TNM~\footnote{A staging system to describe the amount and spread of cancer in a patient’s body, using TNM from AJCC. 
% T describes the size of the tumor and any spread of cancer into nearby tissue; N describes the spread of cancer to nearby lymph nodes; and M describes metastasis (spread of cancer to other parts of the body)
} staging~\citep{amin2017ajcc} combined with pathology report and other multidimensional patient features to perform treatment planning. 
% 近年来的医学视觉–语言模型（VLMs）更多关注于生成全局研究级报告，但在肿瘤领域仍存在重要局限性： 
% \textbf{（i）肿瘤特异性有限。} 现有的VLMs~\citep{bai2024m3d,xin2025med3dvlm,xu2025lingshu}多以疾病分类、视觉测量、病灶检测等简单通用医学任务为主。在放射组学的临床场景中，难以实现病灶roi区域的精准定位，缺乏从影像发现到病理级别风险预测的多模态综合决策能力。
% \textbf{（ii）数据集多样性和专业性不足。} 现有肿瘤数据集数量有限，很少支持对同一病例的多粒度、多角度深度分析，并且数据集大多为知识量较少的选择和短文本问答，缺乏临床问题设计。在器官和肿瘤级别进行临床意义深远的癌症评估时，器官子结构及病灶的细粒度视觉表征常与文本推理错位~\citep{pan2025medvlm}，进一步增加医学结论错误或虚构的风险。 % 
% \textbf{（iii）推理深度不足。} 大部分med-lVLMs~\citep{hamamci2024developing,wu2025towards}大多接收2d格式的医学图像并采用单步推理来完成下游医学任务。而在3D放射组学场景中，图像结构与信息远比2D更为复杂，单步推理方法难以实现多层次、全过程的临床推理。此外，对于如何将影像发现逐步推理至印象，再进一步关联至病理级别诊断和TNM分期，其临床推理链条的构建与评估尚不充分。 
Recent medical LVLMs (Med-LVLMs) primarily focus on study-level report generation. Nevertheless, these models exhibit notable limitations when applied to clinical oncologic scenarios: \textbf{(i)  Limited Tumor-centric Specialization.} Current Med-LVLMs are optimized for broad clinical tasks (report generation, classification, coarse detection)~\citep{bai2024m3d,xin2025med3dvlm,chen2025large}, but fall short in oncology-specific workflows: they do not reliably connect radiologic findings to pathology-level endpoints (TNM stage, nodal metastasis, risk stratification), and thus remain insufficient in clinical oncologic decision-making.
\textbf{(ii) Scarcity of Diverse and Tumor-specific Datasets.} Tumor-related data are scarce and difficult to collect in existing medical datasets such as CT-RATE~\citep{hamamci2024developing} and 3D-RAD~\citep{gai20253d}, which are designed as knowledge-constrained multiple-choice and short-text QA tasks and rarely support multi-granular analysis of individual cases.
To this end, fine-grained visual representations of organ substructures and lesions are often misaligned with textual reasoning~\citep{pan2025medvlm}, increasing the risk of erroneous or fabricated medical conclusions.
\textbf{(iii) Insufficient Reasoning Depth.} Most Med-LVLMs~\citep{qiu2022learning,wu2025towards,xu2025lingshu} are restricted to processing 2D medical images and rely on single-step reasoning for downstream tasks. In 3D radiologic scenarios, the greater structural and informational complexity makes single-step reasoning inadequate for multi-stage clinical inference. Additionly, the construction and assessment of reasoning chains remain insufficient.

 % 为填补上述空白并推动基于影像的肿瘤多模态基础模型的发展，本工作聚焦于五个具有临床意义的消化系统重要器官——食管、胃、结肠、胰腺和肝脏（见图 \ref{fig:intro} (a)），首次提出了以肿瘤为中心贯穿“影像学发现、影像学印象到病理级别诊断”全过程的多模态思维链推理流程（如图 \ref{fig:intro} (b) 所示）。
 % 我们构建了相关CoT数据集并开发了配套的评测基准与3d ct专用模型来评估我们提出多模态推理链条的理论有效性与临床价值，希望能够推动医学AI在肿瘤全流程可解释场景下的创新发展。
 % （i）Dataset. 我们联合多家医疗机构，整合三维CT影像、影像学文本报告和病理报告三类数据，并收集了DeepTumorVQA公开数据集，构建了已知规模最大的多模态肿瘤相关语料库。我们设计了多智能体肿瘤数据生成引擎和五大器官影像诊断指南知识图谱，通过多智能体交互校验，将原始报告转化为四类遵循知识图谱的链式思维（CoT）VQA任务：器官及病灶定位、病灶属性分析、TNM病理预测以及逐步链式报告生成。最终，我们累计构建了超过150万条VQA指令样本的数据集（见图 \ref{fig:intro} (c)）。此外，我们还对DeepTumorVQA公开数据集进行了改写，生成了约35万条CoT风格的数据样本，为医学视觉语言模型的开发与评估提供了更加贴近真实肿瘤决策场景的数据支持。

% （ii）Benchmark. 我们提出了一套面向临床肿瘤链式推理的标准化评测协议，通过从推理链中抽取聚焦于器官、解剖结构和病灶等主体的结构化三元组（Subject, Relation, Object），并借助GPT-4o对每个推理步骤的质量进行评估。该协议不仅关注影像异常特征检测的准确性，还系统考察后续影像学结论与病理预测等推理链环节的逻辑正确性，从而为医学视觉语言模型的临床相关性和综合性能提供了更加全面且权威的评价标准。

%To address these challenges and advance LVLMs for clinical tumor analysis, we focus on five clinically significant digestive organs—the esophagus, stomach, colon, pancreas, and liver (see Fig. \ref{fig:intro} (a)), and propose a tumor-centric multimodal Chain-of-Thought (CoT) reasoning framework that spans the entire clinical process from radiological findings and impressions to pathology-level diagnosis for the first time (see Fig. \ref{fig:intro} (b)). We construct a tumor-specific CoT dataset characterized by multi-granularity, multi-task, and multi-step reasoning, and develop corresponding benchmark protocols and a customized 3D CT reasoning model to systematically evaluate the theoretical validity and clinical utility of the proposed multimodal reasoning framework:

To address these challenges, we build a large-scale, multi-institutional dataset targeting five major digestive organs (Fig.~\ref{fig:intro}a)—and present a tumor‑centric multimodal Chain‑of‑Thought (CoT) framework that operationalizes the full clinical pipeline from radiology findings and impressions to pathology‑level diagnosis (Fig.~\ref{fig:intro}b). 

%\textbf{(i) Dataset.} We collaborate with multiple medical institutions to integrate 3D CT images, radiology reports, and pathology reports, and also incorporate the public DeepTumorVQA~\citep{chen2025vision} dataset, resulting in the largest known multimodal tumor-related corpus. We designed a multi-agent tumor data generation engine and developed knowledge graphs for five major organs. Through multi-agent interactive-validated, original reports were converted into four types of knowledge graph-guided CoT VQA tasks: localization, lesion attribute analysis, TNM pathology prediction, and CoT report generation. As a result, we curate a dataset that contains over 1.4 million VQA instruction samples (see Fig.~\ref{fig:intro} (c)) and approximately 400K CoT-style samples reformulated from DeepTumorVQA, called \texttt{TumorCoT-1.5M}.

\textbf{(i) Dataset.} We collect multi-institutional 3D CT images from patients with pathology-confirmed tumors, each paired with the corresponding clinical radiology and pathology report
%and incorporate DeepTumorVQA~\citep{chen2025vision}
. We also designed a multi‑agent, knowledge‑graph–guided engine that converts original reports into four CoT VQA tasks—localization, lesion attributes, TNM pathology prediction, and CoT report generation—yielding the largest known multimodal tumor-related dataset \texttt{TumorCoT‑1.5M} with approximately 1.5M CoT VQA instruction samples(Fig.~\ref{fig:intro}c).

%\textbf{(ii) Benchmark.} We introduce a standardized protocol for evaluating clinical tumor CoT analysis, which extracts structured subject–relation–object triplets focused on organs, anatomical structures, and lesions from the reasoning chain, e.g. "pancreas-exists-cancer". Each reasoning step is evaluated for quality using GPT-5. The protocol not only measures the accuracy of abnormal feature detection but also assesses the logical correctness of subsequent reasoning steps, including radiological conclusions and pathology predictions, which provides a more comprehensive and authoritative benchmark for evaluating both the clinical relevance TUMOR and overall performance of Med-LVLMs.

\textbf{(ii) Benchmark.} We introduce a standard evaluation protocol that extracts ``subject–relation–object" triplets (e.g. ``pancreatic tail-discovered-malignant tumor") from CoT chains and performs stepwise scoring with an LLM‑based evaluator. It measures the accuracy of abnormality detection and the logical correctness of downstream conclusions—from findings to impressions to pathology—providing a comprehensive assessment of clinical relevance and overall performance of Med-LVLMs.

% （iii）Model. 技术上，我们提出了\textbf{TumorChain}，这是一种具备拓扑感知能力的混合模型交错推理框架，将临床先验和器官拓扑对齐机制注入到视觉—语言推理过程中。考虑到五大消化系统器官的肿瘤往往可发生转移并影响解剖相关区域，TumorChain实现了多轮交错推理：首先，将视觉编码器的特征表示与任务提示输入LLM（大语言模型），以激发对相关器官和区域的关注及“中间思考”；随后，基于LLM输出，由插件式分割专家（TotalSegmentor~\citep{wasserthal2023totalsegmentator}）对指定区域提取ROI级别特征，并将这些特征反馈给LLM，从而生成包含全局影像特征、专家知识文本、局部ROI特征和任务语境的高精度对齐表达。当模型内部推理指示仍有待关注区域时，该交错循环将不断重复。

% 为更好地优化上述架构，我们采用协同的小/大模型联合训练策略：分割专家产生空间器官掩码，局部器官特征配备轻量化分类头以增强视觉编码器对器官层面异常的判别能力，而LLM则负责多模态信息融合与高级临床推理。整体上，这些组件协作实现了基于拓扑的全局—局部特征对齐和有临床基础的推理链，有效降低模型在推理过程中的幻觉风险，并显著提升了影像学发现、影像学印象到病理诊断的预测能力。

\textbf{(iii) Model.} Technically, we propose \textbf{TumorChain}, a topology-aware, hybrid-model interleaved reasoning framework that integrates clinical priors and organ topology. As tumors frequently affect sub-structures and surrounding organs or distantly metastasize, TumorChain performs multi‑round interleaved reasoning: the LLM ingests global tokens and the task prompt to surface intermediate “thoughts” and candidate organs; a segmentation expert returns masks and ROI‑level tokens for the indicated regions; the augmented tokens are fed back to the LLM and the loop continues until no new ROIs emerge, producing a progressively refined CoT chain that fuses global context, local evidence, and task intent. To better optimize this framework, we adopt a collaborative hybrid-model joint training strategy: a segmentation expert produces spatial organ masks, a lightweight abnormality classifier built on local organ features enhances the visual encoder’s organ-level abnormality discrimination, and the LLM performs multimodal integration and high‑level clinical reasoning. This design enables topology‑aware global–local fusion, reduces hallucinations, and improves end‑to‑end tumor assessment from findings to impressions to pathology.

The main contributions of this paper can be summarized as follows:

% \begin{itemize}[leftmargin=1em, itemsep=0em, parsep=0em, topsep=0em]
%     \item \textbf{Clinical Tumor Reasoning Formulation}: We formulate the clinical tumor analysis task into a complete reasoning pipeline from radiology findings → study‑level impressions → pathology predictions. This pipeline not only covers the core oncologic workflow, but also ensures traceability and interpretability, thereby reducing diagnostic errors.
%     \item \textbf{Tumor CoT Data Engine and Evaluation}: We design an interactive-validated data engine and construct a large CoT‑annotated VQA corpus (1.5M instances) for tumor understanding across five digestive organs, together with a CoT evaluation protocol that measures stepwise clinical reasoning.
%     \item \textbf{Interleaved Multimodal Reasoning}: A medical multimodal interleaved‑reasoning framework (\textbf{TumorChain}) with hybrid-model collaborative optimization (i.e. a segmentator, an abnormality classifier, and an LVLM) that achieves organ‑level, global–local multimodal alignment for fine‑grained 3D tumor analysis.
%     \item \textbf{Substantial Performances and Insights}: Extensive experiments across downstream tumor tasks show consistent gains on the \texttt{TumorCoT‑1.5M} test set and strong generalization to the public DeepTumorVQA~\citep{chen2025vision} benchmark. Our work provides actionable insights to develop multimodal foundation models in high‑stakes clinical tumor analysis.
% \end{itemize}

$\bullet$ \textbf{Clinical Tumor Reasoning Formulation}: We formulate the clinical tumor analysis task into a complete reasoning pipeline from radiology findings → study‑level impressions → pathology predictions. This pipeline not only covers the core oncologic workflow, but also ensures traceability and interpretability, thereby reducing diagnostic errors.

$\bullet$ \textbf{Tumor CoT Data Engine and Evaluation}: We design an interactive-validated data engine and construct a large CoT‑annotated VQA corpus (1.5M instances) for tumor understanding across five digestive organs, together with a CoT evaluation protocol that measures stepwise clinical reasoning.

$\bullet$ \textbf{Interleaved Multimodal Reasoning}: A medical multimodal interleaved‑reasoning framework (\textbf{TumorChain}) with hybrid-model collaborative optimization (i.e., a segmentator, an abnormality classifier and an LLM) that achieves organ‑level, global–local multimodal alignment for fine‑grained 3D tumor analysis.

$\bullet$ \textbf{Substantial Performances and Insights}: Extensive experiments across downstream tumor tasks show consistent gains on the \texttt{TumorCoT‑1.5M} test set and strong generalization to the public DeepTumorVQA~\citep{chen2025vision} benchmark. Our work provides actionable insights to develop multimodal foundation models in high‑stakes clinical tumor analysis.

%为了解决上述问题，并推进多模态大模型在基于影像的肿瘤理解，在本文中，我们首先提出了全新的肿瘤多模态任务。与现有的RadFM,lingshu等医疗影像多模态方法不同，如图一（a）所示，我们关注常见的消化系统相关的5大器官的肿瘤：食管、胃、肠、胰腺和肝。如图一（b）所示，我们提出包含完整的影像finding、影像impression到病理结论的临床逻辑链条的肿瘤理解任务。为了训练更好的临床肿瘤VLM，我们利用肿瘤病人影像、影像报告、病理报告，构建了如图一（c）所示的包含Position, lesion attributes, TNM pathology prediction,CoT 分析的报告生成4种类型的VQA任务。共计超过1.5M的VQA instruction。

%此外，我们进一步提出一种临床CoT分析评估的协议，既评估肿瘤异常发现的正确性，同时考虑临床发现和肿瘤临床推理链的正确性，以支持更加全面的评价VLM生成过程的正确性。

%在技术上，我们提出一种引入临床先验知识和器官拓扑结构对齐的视觉-语言跨模态交错推理方法。临床上，消化道5大器官的肿瘤通常会转移进而影响其他关联器官。TumorChain采用多轮交错推理的方式，首先将视觉视觉encoder编码的token和qeustion传入LLM，引导模型思考并关注异常器官相关的其他器官和区域。
%第二步，根据LLM输出的思考信息，利用totalsegmentor专家模型抽取LLM关注区域特征，注入LLM，形成全局影像特征，知识文本，ROI局部影像特征，任务文本这样细腻度交错对齐推理的范式。如果LLM输出还需要关注其他器官或区域进行综合判断，再继续进行上述交错推理范式。
% 为了使得TumorChain更好的优化，我们使用大小模型协同优化的方式，既使用一个分割专家模型负责器官空间分割，一个添加在器官局部特征后面的分类头负责增强视觉encoder对于局部器官异常的鉴别能力，LLM大模型负责多模态信息融合和推理。
%通过这种方式，TumorChain获得基于临床逻辑的跨模态全局、局部特征对齐推理链,自我思考大幅减少包含幻觉的推理结论，提升影像finding-impression-病理的预测能力。

%我们提出了全新的肿瘤理解任务，从影像发现-》影像impression-〉病理预测的肿瘤全流程理解和评估学习范式
%我们根据临床知识设计了肿瘤理解的数据生成引擎，成功构造了150w级别的消化道5癌肿瘤理解VQA数据，并提出具备临床推理逻辑的CoT评估方式。
%我们提出TumorChain，一种大小模型协同优化的多模态交错推理框架，实现临床器官级别的细腻度交错对齐肿瘤推理。
%大量实验验证TumorChain在多个肿瘤理解下游任务上的巨大性能提升，提供多模态大模型在临床重大疾病上的应用insight。

% \begin{wrapfigure}[8]{t}{0.395\textwidth}
% % \vspace{-17mm}
%   \centering
%     \includegraphics[width=0.36\textwidth]{figs/data2.pdf}
%     % \vspace{-3mm}
%   \caption{\texttt{TumorCoT-1.5M} statistics.}
%   \label{fig:cot_data}
% \end{wrapfigure}

\section{TumorCoT-1.5M Dataset}
\subsection{Data Collation and Organization}
\noindent
\begin{minipage}{0.61\textwidth}
Most public medical datasets focus on general healthcare scenarios, featuring simple multiple-choice or short-answer formats lacking stepwise reasoning chains or region-specific causal annotations. To advance reliable tumor analysis with LVLMs, we curate \texttt{TumorCoT-1.5M}, a large-scale dataset centered on comprehensive tumor analysis in 3D CT, encompassing the entire clinical workflow from radiological findings and impressions to pathological prediction.

For data collection, we collaborated with multiple medical institutions to collect 3D CT scans from patients with tumors originated from five major digestive organs (liver, pancreas, stomach, colon, and esophagus). \texttt{TumorCoT-1.5M} includes approximately 41,059 CT images with corresponding captions, 10,708 radiology reports, and partial pathology reports. 

\end{minipage}
\hfill
\vspace{-2mm}
\begin{minipage}{0.36\textwidth}
\centering
\includegraphics[width=\textwidth]{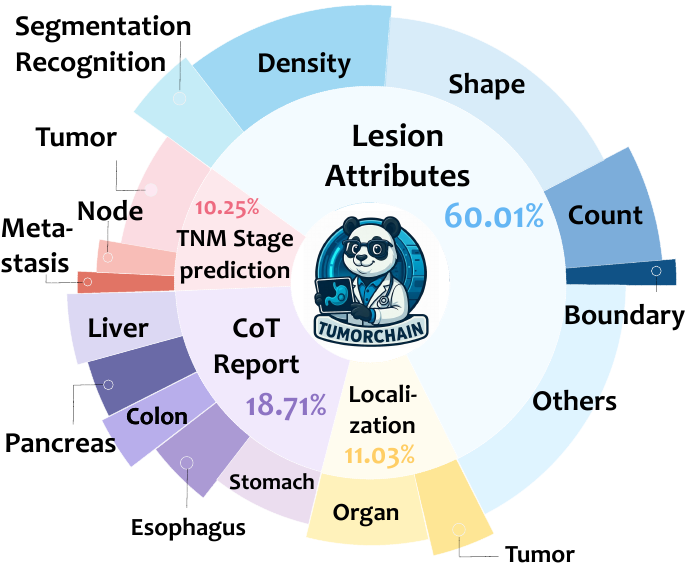}
\captionof{figure}{TumorCoT-1.5M statistics.}
\label{fig:cot_data}
\end{minipage}

% gap-\input{sec/tex/4_Method}》数据来源、数据量-》数据构造方法-〉数据任务类型
% 现有的公开医疗影像数据集大多聚焦于通用医疗保健领域，形式大多为简单的选择题或者短句问答，缺乏临床问题设计和逐步推理链条标注，不支持多步骤、分区域的因果推理过程。为了推动高质量、高可信多模态肿瘤分析lvlms的发展，我们聚焦于ct影像中的癌症发现和风险评估，构建了覆盖从影像发现到放射学impression再到病理预测的肿瘤全流程分析的大型数据集TumorCoT-1.8M和针对可追溯思维链的评价体系TumorChain-Eval。

% 为了提升我们提出的closed-loop multimodal reasoning pipeline的逻辑可靠性和临床预测准确性，我们与多家医疗机构合作收集了消化道五大器官肿瘤检查的CT影像数据。此外，我们还收集了目前最大的肿瘤数据集Deeptumorvqa一起构造了TumorCoT-1.5M。这些数据包含了大约4w张ct影像与对应的caption、1w影像报告以及部分病理报告,涵盖了肝脏、胰腺、胃、结肠、食管消化道五大器官。其中，Deeptumorvqa来源于17个公开数据集，共9262个CT样本，数据包含一些ct影像和病灶的结构化信息。
During the data partitioning stage, we strictly followed a patient-level protocol to divide the dataset into training and test sets at a ratio of 9:1, ensuring that there is no overlap between the two sets and preventing any data leakage.

To fully leverage tumor CT data from diverse sources and formats, we developed an interactive-validated CoT data engine to construct \texttt{TumorCoT-1.5M}. In addition, panels of radiology and pathology experts conducted rigorous reviews of the reasoning chains(details see Appendix~\ref{sec:Formal-evaluation}), supplementing missing links, and ensuring logical consistency and traceability at every step.

% \begin{wrapfigure}[13]{t}{0.4\textwidth}
% % \vspace{-6mm}
%   \centering
%     \includegraphics[width=0.4\textwidth]{figs/data2.pdf}
%     % \vspace{-3mm}
%   \caption{TumorCoT-1.5M Statistics.}
%   \label{fig:cot_data}
% \end{wrapfigure}

Ultimately, as shown in Figure \ref{fig:cot_data},  \texttt{TumorCoT-1.5M} comprises 1,497,818 CoT-VQA pairs, covering four key tasks in tumor analysis: \textbf{(i) Localization}, \textbf{(ii) Lesion Attribute Analysis}, \textbf{(iii) TNM Stage Prediction}, and \textbf{(iv) CoT Report Generation}. Each type is further divided into subtasks based on organ substructures or tumor grades, with tasks ranging from simple to complex in both multiple-choice and open-ended formats. Every sample includes a traceable reasoning process and summary. This organizational framework markedly improves the model’s reasoning interpretability and cross-modal consistency.

 % (b) The overview of the interactive-validated CoT Data Engine, which is a system of multi-agent collaboration, including procedure of raw data processing, diagnostic knowledge graph importing, and CoT data building and validation.
 
% \begin{wrapfigure}[13]{t}{0.47\textwidth}
%     \begin{center}
%     % \vspace{-7mm}
% \includegraphics[width=0.45\textwidth]{figs/fig_data_stat.pdf}
%         \vspace{-3mm}
%         \caption{Data statistics of \Dataset}
%         \label{fig:dataset-stats}
%     \end{center}
% \end{wrapfigure}
\subsection{CoT Data Engine}
% 我们设计了一个交错校验的multi-agent数据引擎来整合放射科医生和病理专家的专业知识，构造了TumorCoT-1.5M。我们收集的每条案例均包含病灶区标注与影像描述，数据引擎依托专业医生制定的知识库在肿瘤分析的ROI定位、CT理解和病理预测三大步骤上构造不同细粒度的cot vqa。通过多agent的交错校验和知识图谱的引导，原始报告中的放射征象、影像印象与最终病理结果能够实现多层次跨模态对齐。As shown in Figure 3, the agent engine comprises 6 components as follows.

 % We designed an interactive-validated CoT data engine to construct \texttt{TumorCoT-1.5M}. Each case includes lesion annotations and imaging descriptions, and the engine leverages a physician-curated knowledge base to build fine-grained CoT VQA tasks across three major steps: ROI localization, CT analysis, and pathological prediction. Through iterative validation of multi-agent and knowledge graph guidance, radiological findings, impressions, and final pathology outcomes from the original reports are aligned across multiple modalities. As shown in Figure \ref{fig:agent} (b), the data engine comprises the following 6 components:

 We designed an interleaved-validation CoT data engine to construct \texttt{TumorCoT-1.5M}. Guided by an expert-level knowledge graph, we generated diverse fine-grained CoT-VQA samples across three major steps in tumor analysis: ROI localization, CT interpretation, and pathology prediction. As shown in Figure \ref{fig:agent}, the data engine comprises the following 3 procedures with 6 agentic experts (detailed in Appendix~\ref{sec:appendix-data_engine} and details of the 6 experts are in Appendix~\ref{sec:appendix-agents}):
 
% 1）影像诊断知识库。我们联合五名五大器官领域的专业医生，将相关影像诊断指南、专业教材和典型病例构建为“主体—关系—客体”的三元组知识图谱，覆盖器官、解剖结构、影像发现、影像印象、组织病理、风险因素等核心概念(见附录图～)。所有器官分段标准均依照国际主流规范制定，便于后续进行子结构及肿瘤级别的分层分析和推理问答。推理链构建过程中，系统会从知识库检索相关节点和关系，并与原始报告信息共同输入推理模型，确保逻辑链条的可追溯性与高可靠性，减少虚假事实和逻辑错误。

\begin{figure*}[t]
  \centering
  \includegraphics[width=1\linewidth,clip=false, trim=0 15 5 0]{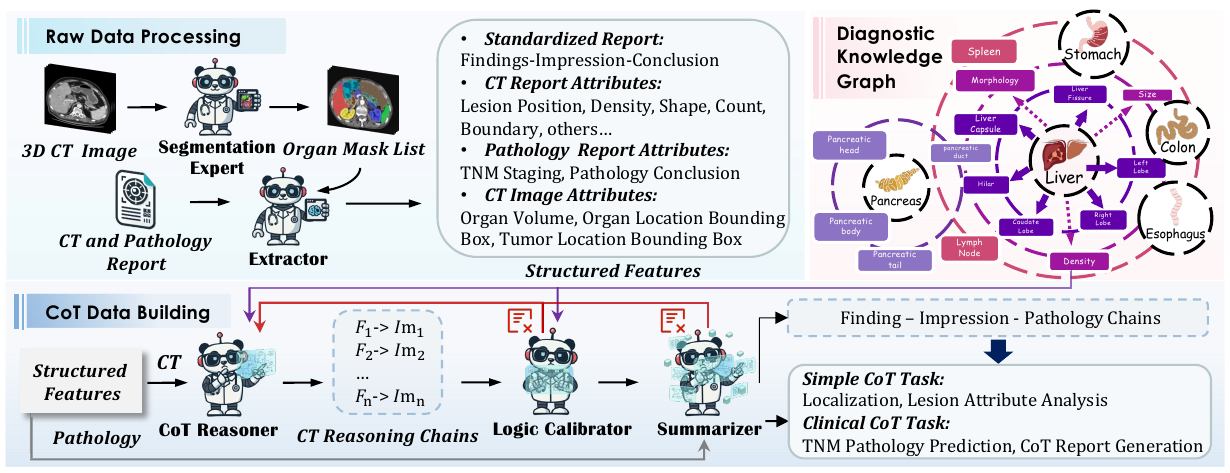}
    \vspace{2mm}
    \caption{The overview of the interactive-validated CoT Data Engine, which is a system of multi-agent collaboration, including the procedure of raw data processing, diagnostic knowledge graph importing, and CoT data building and validation.}
    % \vspace{-4mm}
  \label{fig:agent}
\end{figure*}

\textbf{(i) Raw Data Processing}:
This process includes two expert agents: the Segmentation Expert Model (TotalSegmentator~\cite{wasserthal2023totalsegmentator}) and the Structured Feature Extractor (Qwen3-235B-A22B~\citep{yang2025qwen3}), aiming to clean data, standardize terminology, obtain organ masks(Appendix~\ref{sec:appendix-organ}), and extract structured features from CT and pathology reports.

\textbf{(ii) Diagnostic Knowledge Graph-Driven Prompt Engineering}:
In collaboration with radiologists and pathologists, we construct diagnostic knowledge graphs (KGs) covering five major organs (The KG for 5 digestive organs is shown in Appendix~\ref{sec:appendix-kg}). During reasoning chain construction, relevant nodes and relationships are retrieved and then authoritative clinical guidelines are integrated into the LLM prompts to structurally constrain the model to follow professional medical standards, thereby ensuring traceability and high reliability of the logical chain.

\textbf{(iii) CoT Data Building and Cross-Review Mechanism}:
This process includes 3 agents: \textbf{CoT Reasoner}, \textbf{Logic Calibrator}, and \textbf{Summarizer}. To fully utilize the strengths of different LLMs, three agents are assigned distinct models. Additionally, we use only the textual content from the reports as their sole source of information, thereby preventing these models from making subjective assumptions about complex modalities such as CT imaging. CoT Reasoner uses GPT-4o-mini for its strong language and medical capabilities, processing structured CT report features and integrating external KGs to link radiological findings with impressions. Logic Calibrator, based on Claude3.5-Haiku, automatically validates reasoning accuracy. If potential issues are detected, the system randomly applies two prompting strategies—expanding the organ region or providing a suspected cause—to guide the reasoning model in re-evaluating its chain. Summarizer employs GPT-5-mini to process structured pathology reports, aggregate and verify reasoning chains, and generate QA pairs and CoT-formatted reports when chains align with pathological conclusions; otherwise, it triggers upstream re-reasoning. The agents' prompts are shown in Appendix~\ref{sec:appendix-prompt}.

\subsection{Clinical CoT Evaluation: TumorChain-Eval}
\label{sec:eval_metrics}
%from zimu
%现有的report生成的评估方法大多计算模型输出与gt报告之间的相似度，或者直接采用大模型进行评分。对于医学逻辑推理任务来说，这忽略了临床逻辑的正确性以及对CoT推理过程进行step-wise的评估。为此，我们针对肿瘤临床CoT提出一种新的评估方式。具体地，将GT和pred的CoT推理过程分别提取成三种推理链,可统一表示为subject-relation-object：1) Finding Chain (FC); 2) Impression Chain (IC);3) Long Reasoning Chain (LRC). 
%FC包含影像学初步事实描述，独立存在，不涉及高阶推理，例如 `胃壁-》发现-〉增厚`，肝-》发现-〉高密度影。
%IC 描述基于 Finding Chain 总结的中级医学印象，体现了基于多个 Finding 的简单推论总结。例如 肝肿块/肝高密度影-提示-局部病理性改变`。
%LRC综合多个 Finding 和 Impression，生成更高层级的医学推理（如 肝肿块/高密度影/累计脾脏`-》高度怀疑-〉肿瘤性病变`，结论性提示具有较强的逻辑推导性。
%FC、IC和LRC每一项都具有细化的单项评分规则，采用GPT4o模型根据评分规则进行打分，最后根据各项的权重计算最后的CoT-e score.

Existing evaluation metrics like accuracy, semantic similarity metric (Rouge-L~\citep{lin2004rouge}, CIDEr~\cite{vedantam2015cider}, RaTEScore~\cite{zhao2024ratescore}, etc.) ignore the correctness of clinical logic and step-wise evaluation of the CoT reasoning process. 
To address this issue, we propose \textbf{TumorChain-Eval}, a novel evaluation framework with metric $CoT_e$ specifically for oncology-related CoT reasoning, which provides a fine-grained and interpretable scoring system based on reasoning logic chains.
In this approach, the CoT reasoning process of the ground truth and prediction is extracted into three reasoning chains, all of which can be represented as subject-relation-object triplets:
\textbf{(i) Finding Chain}: Describes the primary radiology findings, which are independent facts and do not involve higher-order reasoning. 
% Example: (stomach wall -- discovered -- thickening) or (liver -- discovered -- high-density shadow);
\textbf{(ii) Impression Chain}: Summarizes intermediate-level clinical impressions based on the Finding Chain, involving simple reasoning conclusions from multiple findings. 
% Example: (hepatic lesion/high-density shadow -- suggests -- localized pathological changes).
\textbf{(iii) Long Reasoning Chain}: Combines multiple findings and impressions to generate higher-order reasoning, such as conclusive diagnostic hypotheses with strong logical inference. 
% Example: (hepatic lesion/high-density shadow/infiltration of spleen -- highly suspicious of -- malignant tumor).
Each type of chain is evaluated using a detailed set of scoring criteria, and the scores are calculated by using GPT-4 with scoring guidelines. Finally, the $CoT_e$ score is computed by weighted sum of $S_{FC}$, $S_{IC}$, $S_{LRC}$, which are the FC score, LC score, and LRC score, respectively.
% \begin{equation}
% CoT_e = W_{FC} \cdot \frac{1}{N}\sum^N_0(S^i_{FC}) 
%                + W_{IC} \cdot \frac{1}{N}\sum^N_0(S^i_{IC}) 
%                + W_{LRC} \cdot \frac{1}{N}\sum^N_0(S^i_{LRC}) ,
% \end{equation}
% where $S^i_{FC}$ is the score the sample $i$ in FC scoring, $S^i_{IC}$ is the score the sample $i$ in FC scoring, $S^i_{LRC}$ is the score the sample $i$ in FC scoring, $N$ represents sample number, and $W$ denotes weighting coefficients.
% where weight $W$ denotes the relative importance of each reasoning chain, and can be adjusted according to their specific clinical relevance in the evaluation context.
The detailed scoring process and criteria for each reasoning chain can be found in the Appendix~\ref{sec:appendix-cot_eval}.

\section{Methodology}
\subsection{Overview of TumorChain}
\begin{figure*}[t]
  \centering
  \includegraphics[width=1\linewidth,clip=false, trim=0 8 5 0]{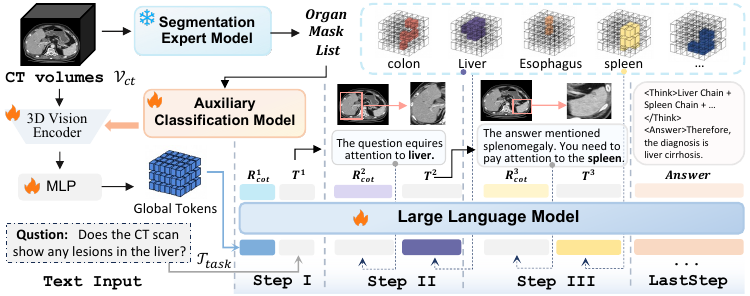}
      \vspace{2mm}
    \caption{The overview of \textbf{TumorChain}.
    Given CT volumes and task prompts, TumorChain analyzes the task requirements and performs iterative interleaved reasoning by integrating a segmentation expert model for ROI localization, auxiliary classification for local feature enhancement, and organ-guided iterative reasoning for clinical tumor analysis.
    }
  \label{fig:model}
\end{figure*}

Before presenting our method, we first introduce some
basic notions and terminologies.
The framework $\mathcal{F}(\cdot)$ of TumorChain is shown in Figure \ref{fig:model}, which consists of five key modules: a 3D vision encoder $\mathcal{{E}}_{v}(\cdot)$, an organ segmentation expert $\mathcal{S}eg(\cdot)$, an auxiliary classification model $\mathcal{C}ls(\cdot)$, a multi-layer perceptron (MLP) projector 
 $\mathcal{P}(\cdot)$, and a LLM $\mathcal{LLM}(\cdot)$. Given the CT volumes $\mathcal{V}_{ct} \in \mathbb{R}^{H\times W\times D}$ with task prompt text $\mathcal{T}_{task}$ as input, the TumorChain outputs the response $\mathcal{R}_{cot}$ of the interleaved multi-modal chain of thought (CoT), which simplifies as below:
 \begin{equation}
\label{eq:llm_out0}
  \underbrace{\mathcal{F}(\mathcal{V}_{ct}, \mathcal{T}_{task})}_{\rm \textbf{TumorChain}} : \underbrace{\mathcal{{E}}_{v}(\cdot) \rightarrow \mathcal{S}eg(\cdot) \rightarrow \mathcal{C}ls(\cdot)\rightarrow \mathcal{P}(\cdot)}_{\rm {\textbf{Global\&Local \, Visual \, Alignment}}} \rightarrow \!\!\!\!\!\!\!\!\!\!\underbrace{\mathcal{LLM}(\cdot) \rightarrow \mathcal{R}_{cot} }_{\rm \textbf{Interleaved \, Multi-modal \, Inference}}
\end{equation}
 
% \textbf{Step I:  Global\&Local Visual Alignment.}
\textbf{Global \& Local Visual Alignment.}
For global visual tokens, the 3D encoder $\mathcal{E}_v$ takes the $V_{ct}\in \mathbb{R}^{H\times W\times D}$ as input that encode 3D CT volumes to a series of vision tokens as $\tau_v = \mathcal{E}_v(V_{ct})$. 
Then, all vision tokens are aligned into LLM space by the projector $\tau_{g} = \mathcal{P}(\tau_v)$, where $\tau_g$ denotes the global tokens and has a size of $L_g\times K$ ($L_g$ tokens and embedding dimension K).

For local visual tokens, the organ segmentation expert $\mathcal{S}eg(\cdot)$ first segments the organ mask as $\mathcal{M}_{organ} = \mathcal{S}eg(V) \in \mathbb{R}^{H\times W\times D}$ to provide fine-grained location for interlaced reasoning. According to the task prompt, TumorChain identifies the target task organ mask $\mathcal{M}_{task}$ from $\mathcal{M}_{organ}$ by matching the organ in the task text and extracts local vision tokens $\tau_l$ as
\begin{equation}
\label{eq:mask}
    \tau_l = \Gamma(\tau_g, \mathcal{M}_{task}) \quad s.t. \quad \mathcal{M}_{task} = \Lambda(\mathcal{M}_{organ}, \mathcal{T}_{task}),
\end{equation}
where $\Lambda(\cdot)$ denotes the organ matching operation and $\Gamma(\cdot)$ represents the operation of extracting local token $\tau_{l}$ of size $L_l\times K$ according to mask.

Besides, to enhance the $\tau_l$ learning, we introduce the auxiliary classification model $\mathcal{C}ls(\cdot)$ that projecting $\tau_l$ into two classification logits, which can be formulated as ${y} = Cls(\tau_l)$, where the  $\mathcal{C}ls(\cdot)$ distinguish normal or abnormal for each local organ, thereby significantly enhancing the encoder's discriminative capability. 
When the visual features and task prompts are jointly input into the LLM, the model’s attention to key information and anomalous regions can be further enhanced. 

\textbf{Interleaved Multi-modal Inference.}
% \textbf{Step II: Interleaved Multi-modal Inference.}
After obtaining all vision and text tokens, they are combined into interleaved tokens to serve as the input for LLM, to further generate the inference response $\mathcal{R}_{cot}$ of CoT as 
\begin{equation}
\label{eq:llm_out}
    \mathcal{R}_{cot} = \mathcal{LLM}(\tau_{in}) \quad s.t. \quad \tau_{in} = [\tau_g, \mathcal{T}_{task}, \mathcal{T}^1, \tau_{l}^1, \mathcal{T}^i, \tau_l^i,...],
\end{equation}
where $\tau_{in}$ indicates all input tokens of LLM, and is the combination of global vision tokens $\tau_g$, task text prompt tokens $\mathcal{T}_{task}$, and task target organ tokens $\tau_l^i$ with its text pair $\mathcal{T}^i$. Here $i \in \{1,\dots,\mathcal{O}\}$ indexes the $\mathcal{O}$ potential related or ROI organs/suborgans reasoned by LLM.  

Through the aforementioned process, TumorChain achieves interleaved alignment of global visual features, task text, and task-specific local visual features during a single inference by leveraging the collaborative interaction between large and small models.

\subsection{Organ-guided Iterative Interleaved Reasoning}
%Clinical 3D tumor analysis requires identifying organs within large-scale human body scans, further localizing tiny tumors, and providing sufficiently reliable evidence to support LLM-based clinical reasoning. This makes single-pass inference inadequate and poses challenges for multimodal evidence-based self-validation.
Clinical 3D tumor analysis demands organ identification, fine‑grained tumor localization, and reliable ROI evidence for LLM reasoning, making single‑pass inference insufficient and complicating multimodal self‑validation.
To achieve a reliable clinical reasoning chain of ``finding → impression → pathology prediction", TumorChain adopts an \textbf{organ-guided Iterative Interleaved Reasoning (IIR)}.
%The key insight of IIR lies in leveraging the iterative outputs of the LLM to refine and identify more granular visual ROI regions, which are interleaved with the LLM outputs to serve as inputs for the next iteration. This enables multi-round interleaved reasoning while simultaneously performing self-validation to ensure alignment between visual features and textual outputs across iterations.
IIR uses each LLM output to refine ROI selection, feeds the resulting ROI features back for the next round, and repeats, which enables multi‑round interleaved reasoning with built‑in self‑validation that maintains visual–text alignment across iterations.
For clarity, we abstract the IIR process into three steps:
\paragraph{Step I: Initial Round of Reasoning.}
The initial reasoning step takes the encoded global CT volume tokens $\tau_g$, and the medical question text $T_{task}$, as the inputs to the LLM:
\begin{equation}
    \mathcal{R}^1_{cot} = \mathcal{LLM}(\tau_g, T_{task}),
\end{equation}
where $\mathcal{R}^1_{cot}$ represents the initial diagnostic result output by the LLM.

\paragraph{Step II: Interleaved Self-reflection and Organ Localization.}
\textit{(i) Keyword Extraction and Organ Identification:} According to the initial output $R^1_{cot}$ and Equation \ref{eq:mask}, determining the local ROI mask $M_l^1$ to obtain local vision tokens $\tau_l^1$.
\textit{(ii) Augmented Prompt Construction:} After generating organ-specific tokens based on the extracted organ features, we then construct augmented prompts as:
\begin{equation}
    \mathcal{T}^1_l = (\text{``\texttt{The question requires greater attention to}"}, organ^1_l),
\end{equation}
where $organ^1_l$ represents the identified ROI from $R^1_{cot}$.
\textit{(iii) Combining Inputs for the Next Iteration:} Then, the original question $\mathcal{T}_{task}$, the initial answer $R^1_{cot}$, and identified organ tokens $\tau^1_l$ are combined to form a new input for the LLM:
\begin{equation}
\label{eq:ll1}
    \tau_{in}^2 = [\tau_g, \mathcal{T}_{task}, \mathcal{R}^1_{cot}, \mathcal{T}^1_l, \tau^1_l].
\end{equation}

\paragraph{Step III: Iterative Causal Reasoning}
After completing the previous self-reflection, the framework continues to perform iterative reasoning based on the newly generated answers. If additional relevant organs are introduced, the ROI extraction process is repeated to sequentially obtain the features of all organs involved in the reasoning chain, supporting further verification.
According to Equation~\ref{eq:llm_out} and \ref{eq:ll1}, this iterative process can be formulated as
\begin{equation}
    % \mathcal{R}^{i+1}_ = LLM(\tau_g, T_{task}, R^1, T^1_l, \tau_l^{i}, ..., R^{i}, T^{i}_l,\tau^i_l).
    \mathcal{R}^{i+1}_{cot} = \mathcal{LLM}(\tau_g, \mathcal{T}_{task},  \mathcal{R}^{i}_{cot}, \mathcal{T}^i, \tau_l^{i}).
\end{equation}

For example, in cases where cirrhosis of the liver is suspected and the imaging findings indicate splenomegaly and altered hepatic lobe proportions, IIR automatically extracts the organ tokens for the spleen and appends a pertinent prompt before it feeds both the original question and the previous answers into the LLM. The framework undertakes multi-round causal verification of each organ in the answer until all relevant organs have undergone feature extraction and causal reflection.

\subsection{Hybrid-model Collaborative Optimization Strategy}
Typically, a 3D CT scan contains over 100 slices, each with a resolution of $512\times 512$ pixels. In comparison to 2D image-caption pairs (e.g., resolutions of 224 or 384), this makes it challenging to achieve fine-grained alignment between the visual representations and textual descriptions.
To address the above challenges and further optimize TumorChain, we employ a Hybrid-model Collaborative Optimization (HCO) strategy. The core concept of HCO lies in the collaboration of models at different scales: the segmentation model localizes ROIs, continuously providing precise local features for reasoning. The  classification model ensures discriminative local feature learning during optimization, enabling the visual encoder to effectively distinguish abnormal from normal patterns, preventing subtle anomalies from being overshadowed during LLM training. The LLM integrates reasoning outcomes and leverages the segmentation model for iterative decision-making.

% In a nutshell, during training, the 3D CT encoder extracts features, the segmentation model localizes regional features, the classification model enhances the discriminative power of local features, and the LLM integrates global-local visual features with textual knowledge for learning. 
During training, the classification model and LLM are jointly optimized via loss functions.
The training loss $L_{total}$ is:
\begin{equation}
    L_{total} = -\frac{1}{N}\sum_{n=0}^NlogP(r_n|\tau_{in},r_1, ..., r_{n-1}) - \alpha \cdot \frac{1}{M}\sum_{m=0}^M\hat{y}P(y|\tau_l),
\end{equation}
where $N$ represents the output text length and $M$ represents the sample number. $r_i \in R^{N}$.
$\alpha$ is a loss weight. $y$ and $\hat{y}$ are the prediction and ground-truth of classification (normal/abnormal) for local organs.
During inference, the segmentation model utilizes the ROI organ locations provided by the LLM to extract ROI visual features, which are then fed back into the LLM for self-verification. 
\section{Experiments}
\subsection{Experimental Settings}
\textbf{Model Overview.} Both TumorChain-3B and TumorChain-7B employ the pretrained LLM backbone of Qwen2.5-VL-3B/7B as the large language model component, integrating M3D as the vision encoder~\citep{bai2024m3d} to enable 3D CT feature extraction. Full architectural details and training configuration are provided in Appendix~\ref{sec:appendix-training_details}.

\textbf{Baselines.}
To comprehensively evaluate the performance of TumorChain, we
compare it against a diverse set of baseline models, including seven open-world LVLMs 
(e.g., Claude 3~\citep{claude3}, Gemini 2.0~\citep{deepmind_gemini25_report_2025}, GPT-5~\citep{openai_gpt5_system_card_2025}, Qwen2.5-VL~\citep{bai2025qwen2}, InternVL-2.5~\citep{chen2024expanding}, MiniCPM-V4.5~\citep{yu2025minicpmv45cookingefficient}, LLaVA-CoT~\citep{xu2025llava}), three 2D-based Med-LVLMs (e.g., HealthGPT~\citep{lin2025healthgpt}, Lingshu~\citep{xu2025lingshu}, MedVLM-R1~\citep{pan2025medvlm}), and two 3D-based Med-LVLMs (
e.g., RadFM~\citep{wu2025towards}, M3D-Phi-3~\citep{bai2024m3d}).
% Training details are shown in Appendix \ref{sec:appendix-training_details}.
% \textbf{Data Details.}

\textbf{Benchmarks.}
All models are evaluated on the test subset of TumorCoT-1.5M, partitioned via a 9:1 split. %Specifically, for the public DeeptumorVQA benchmark, we adhere to its original setup, including the four designated task types and official train/test splits, to prevent data leakage and maintain fairness. 
Evaluation is conducted based on our definitions of four main tasks and twelve subtask types.

\textbf{Evaluation Metrics.} We developed diverse protocols focusing on the analytical process and conclusions within the reasoning chain. All models are guided to generate CoT reasoning chain, which is scored using our proposed TumorChain-Eval. For conclusion, accuracy is calculated for multiple-choice tasks (e.g., segment localization in Task 2 and Task 1). For open-ended QA tasks, semantic consistency of the conclusions is assessed using GPT-5 and semantic accuracy is computed. Our evaluation prompt design and implementation details are provided in Appendix~\ref{sec:appendix-implementation}.

\subsection{Main Results}
\textbf{CoT-Conclusion Analysis.}
\begin{table*}[t]
\vspace{-0.5cm}
\caption{Comparison of TumorChain with other LVLMs on \texttt{TumorCoT} benchmark. \textbf{Bold} and \underline{underlined} text indicates the best and second-best performance, respectively.}
\vspace{-1mm}
\label{tab:tumor_bench}
\begin{center}
  \renewcommand{\arraystretch}{1}
\resizebox{\linewidth}{!}{
\begin{tabular}{lccccccccccccc}
\toprule
\multirow{3}{*}{\textbf{Model}} 
& \multicolumn{2}{c}{\textbf{Position}} 
& \multicolumn{6}{c}{\textbf{Lesion Attributes}} 
& \multicolumn{3}{c}{\textbf{TNM Prediction}} 
& \multirow{3}{*}{\textbf{\begin{tabular}{@{}c@{}}CoT-\\Report\end{tabular} }}
& \multirow{3}{*}{\textbf{Avg.}} \\
\cmidrule(rl){2-3} \cmidrule(rl){4-9} \cmidrule(rl){10-12}
& \begin{tabular}{@{}c@{}}Organ\\Pos.\end{tabular} 
& \begin{tabular}{@{}c@{}}Tumor\\Pos.\end{tabular} 
& \begin{tabular}{@{}c@{}}Seg.\\Loc.\end{tabular} 
& Shape & Boundary & Density & Count & Others 
& Tumor & Node & Met. & \\
\midrule
\rowcolor{gray!5}
\multicolumn{14}{c}{\textit{Commercial Models}} \\
\midrule
Claude3-Haiku   & 32.61 & 33.05 & 52.38 & 58.51 & 50.00 & 50.74 & 37.51 & 50.95 & 59.19 & \underline{56.59} & 43.40 & 33.21 & 46.51 \\
Gemini2.0-Flash & 43.27 & 17.69 & 46.81 & 35.97 & 38.22 & 39.11 & 40.22 & 52.86 & 34.03 & 42.21 & 49.06 & 56.08 & 41.29 \\
GPT-5-Mini      & 44.18 & 37.73 & 44.96 & 47.72 & 62.58 & 57.13 & 64.85 & 66.67 & 34.25 & 47.40 & 49.60 & 62.02 & 51.59 \\
\midrule
\rowcolor{gray!5}
\multicolumn{14}{c}{\textit{Generalist Models}} \\
\midrule
Qwen2.5-VL-7B   & 29.57 & 32.52 & 27.41 & 43.96 & 57.51 & 49.36 & 46.73 & 58.02 & 34.17 & 37.49 & 51.15 & 60.54 & 44.04 \\
InternVL-2.5-8B & 29.10 & 27.67 & 33.50 & 49.96 & 51.45 & 49.48 & 42.68 & 51.70 & 36.16 & 35.07 & 48.09 & 52.18 & 42.25 \\
MiniCPM-V4.5-9B & 35.70 & 29.56 & 44.11 & 52.61 & 53.67 & 50.80 & 46.92 & 58.45 & 43.12 & 44.74 & 44.27 & 47.84 & 45.98 \\
LLaVA-CoT-11B & 31.63 & 27.64 & 37.70 & 59.44 & 49.34 & 49.60 & 37.90 & 55.48 & 48.20 & 47.91 & 49.23 & 44.81 & 44.91 \\
\midrule
\rowcolor{gray!5}
\multicolumn{14}{c}{\textit{2D Medical Models}} \\
\midrule
HealthGPT-3.8B & 30.58 & 22.80 & 45.87 & 51.35 & 45.77 & 47.27 & 31.24 & 40.13 & 45.68 & 49.02 & 39.69 & 39.20 & 40.72 \\
Lingshu-7B     & 25.28 & 32.58 & 24.90 & 51.70 & 60.61 & 53.40 & 45.21 & 63.63 & 30.35 & 42.79 & \underline{56.49} & 54.30 & 45.10 \\
MedVLM-R1-2B   & 32.23 & 38.92 & 19.37 & 48.79 & 51.20 & 48.02 & 44.53 & 52.26 & 30.38 & 27.26 & 32.82 & 39.70 & 38.79 \\
\midrule
\rowcolor{gray!5}
\multicolumn{14}{c}{\textit{3D Medical Models}} \\
\midrule
RadFM        & 13.86 & 15.97 & 15.23 & 21.25 & 49.84 & 21.07 & 19.34 & 24.15 & 49.58 & 46.02 & 28.47 & 21.34 & 26.32 \\
M3D-Phi-3-4B  & 32.39 & 28.28 & 22.69 & 23.93 & 42.74 & 30.49 & 27.19 & 52.35 & 44.82 & 30.80 & 35.86 & 34.79 & 32.84 \\
\midrule
\rowcolor{gray!5}
\multicolumn{14}{c}{\textit{Our Models}} \\
\midrule
\rowcolor{blue!5}
\textbf{TumorChain-3B} & \underline{99.93} & \underline{96.44} & \underline{79.13} & \underline{69.07} & \underline{73.57} & \underline{68.97} & \underline{81.04} & \underline{76.34} & \underline{86.34} & 55.09 & \underline{56.49} & \underline{78.96} & \underline{76.78} \\
\rowcolor{blue!5}
\textbf{TumorChain-7B} & \textbf{99.97} & \textbf{97.57} & \textbf{86.88} & \textbf{82.28} & \textbf{84.52} & \textbf{85.05} & \textbf{86.20} & \textbf{86.57} & \textbf{88.83} & \textbf{61.63} & \textbf{71.07} & \textbf{82.36} & \textbf{84.41} \\
\rowcolor{violet!5} \quad \textit{w/o} CoT, IIR & 99.92 & 97.45 & 78.95 & 75.78 & 74.60 & 80.18 & 82.82 & 78.75 & 84.37 & 59.29 & 66.85 & 79.83 & 79.90 \\
\rowcolor{violet!5} \quad \textit{w/o} CoT & 99.91 & 97.10 & 82.75 & 78.43 & 82.57 & 82.79 & 84.26 & 84.16 & 88.34 & 60.76 & 68.12 & 80.26 & 82.45 \\
\rowcolor{violet!5} \quad \textit{w/o} IIR & 99.87 & 97.55 & 79.36 & 79.52 & 74.62 & 81.46 & 80.60 & 79.20 & 84.65 & 60.32 & 65.87 & 81.06 & 80.34 \\
\bottomrule
\end{tabular}
}
\end{center}
% \vspace{0.7cm
\end{table*}

% 在我们提出的四类肿瘤分析任务上分别对分析过程和结论两部分进行了评测，其中对于结论的正确性评估的实验结果见表1。主要的observations are as follows:
% 1、 Superior Performance: TumorChain-7B achieves SoTA performance across all subtasks，以平均分84.41%显著领先于所有基线模型，TumorChain-3B同样表现优异，多项任务准确率远超商用模型、通用模型及现有医学视觉语言模型，验证了我们的数据集和模型设计对与复杂肿瘤分析任务的显著效益。
% 2、现有医疗模型的限制。从结果可以看出，当前主流医疗模型在肿瘤分析任务中的表现不仅未能超越通用开源或商用模型，部分指标甚至略有落后。这一现象说明，现有medlvlms普遍依赖经验性的数据驱动方法，在应对复杂专病场景时，泛化与适应能力尚有不足。值得注意的是，部分三维医学模型在三维CT影像分析任务中的表现显著低于二维医疗图像模型，这进一步揭示了当前三维架构在large-scale人体CTscans的空间定位、语义理解和多属性分析方面存在明显瓶颈。
% 3、多细粒度任务设计：比较我们自己的模型在所有子任务上面的实验结果，充分证明了数据集设计的多层次多细粒度特征。其中，定位任务作为最基础和结构化的任务，模型能够快速收敛并取得几乎满分的准确率。而在挑战性的病理预测及报告生成等任务中，模型分数则明显降低，反映出这些任务在肿瘤诊断链条中对模型推理与理解能力的更高要求。结果表明我们的数据设计思路，覆盖了肿瘤分析的关键环节，也为人工智能在医学领域的细粒度认知提供了重要参考价值。
We evaluated both the reasoning process and the correctness of conclusions for our four proposed tumor analysis tasks. The experimental results for conclusion accuracy are presented in Table \ref{tab:tumor_bench}, lines 1 to 13, and organ-level accuracy is displayed in Figure~\ref{fig:case}b. The main observations are as follows: \textbf{(i) Superior Performance.} TumorChain-7B achieves state-of-the-art results across all subtasks, with an average accuracy of 84.41\%, substantially outperforming all baseline models and. TumorChain-3B also demonstrates strong performance, highlighting the effectiveness of our dataset and model design in complex tumor analysis scenarios. \textbf{(ii) Limitations of Existing Med-LVLMs.} Current mainstream 2D or 3D medical models do not surpass, and in some metrics even underperform, compared to general-purpose or commercial models in tumor analysis. This indicates the limited generalization and adaptability of existing Med-LVLMs, especially in complex, disease-specific settings, and in unseen datasets. Notably, some 3D medical models perform significantly worse than 2D models on CT analysis, revealing critical bottlenecks in spatial understanding and multi-attribute analysis of large-scale CT scans. \textbf{(iii) Effectiveness of Multi-granularity Benchmark Design.} The performance of our models across various subtasks demonstrates the multi-level and fine-grained nature of our dataset. Near-perfect accuracy in basic localization tasks and relatively lower scores in more challenging pathological prediction and report generation emphasize the comprehensive and progressive challenges posed by our benchmark, providing valuable insights for fine-grained AI development in the domain of tumor analysis and diagnosis.

\begin{wraptable}[14]{r}{0.43\textwidth} 
    \centering
    \vspace{-7mm}
    \caption{TumorChain-Eval with other \textbf{reasoning LVLMs} on \texttt{TumorCoT}.}
    \vspace{2mm}
    \renewcommand\tabcolsep{9pt}
    \resizebox{0.43\textwidth}{!}{%
    \begin{tabular}{lcccc}
        \toprule
        \textbf{Reasoning Models} & $S_{FC}$ & $S_{IC}$ & $S_{LRC}$ & $CoT_{e}$ \\
        \midrule
        \rowcolor{gray!5}
        \multicolumn{5}{c}{\textit{Commercial Models}} \\
        \midrule
        Claude3-Haiku     & 59.31 & 43.66 & 30.80 & 43.21  \\
        Gemini2.0-Flash    & 63.10 & 57.69 & 45.12 & 54.28  \\
        GPT-5-Mini         & \textbf{64.22} & \textbf{66.42} & \textbf{55.09} & \textbf{61.23}  \\
        \midrule
        \rowcolor{gray!5}
        \multicolumn{5}{c}{\textit{Open-Source Models}} \\
        \midrule
        Qwen2.5-VL-7B      & 49.15 & 22.89 & 12.09 & 26.45 \\
        InternVL-2.5-8B    & 56.66 & 57.92 & 47.05 & 53.19  \\
        MiniCPM-V4.5-9B   & 60.51 & 58.52 & 45.44 & 53.89 \\
        % LLaVA-CoT-11B    & 42.27 & 34.36 & 18.87 & 30.54 \\
        \midrule
        HealthGPT-3.8B     & 49.64 & 49.68 & 37.15 & 44.65 \\
        Lingshu-7B         & 52.51 & 31.93 & 16.46 & 31.92 \\
        MedVLM-R1-2B       & 40.25 & 36.05 & 21.23 & 31.38 \\
        \midrule
        \rowcolor{blue!5}
        \textbf{TumorChain-3B} & 61.75 & 49.99 & 35.68 & 47.98 \\
        \rowcolor{blue!5}
        \textbf{TumorChain-7B} & \textbf{62.37} & \textbf{62.60} & \textbf{52.57} & \textbf{58.33} \\
        \bottomrule
    \end{tabular}
    }
    \label{tab:cot_eval2}
\end{wraptable}

\textbf{CoT-Reasoning Analysis.}
We compare CoT reasoning ability with other reasoning models in Table \ref{tab:cot_eval2} via the proposed $CoT_e$ metric (Section~\ref{sec:eval_metrics}).
TumorChain-7B demonstrates high accuracy in generating traceable reasoning (FC, IC) and complex reasoning chains (LRC), achieving $CoT_e$ scores of 58.33 (see Figure~\ref{fig:case}c for example). Its clinical reasoning ability outperforms the general LVLMs (e.g., Qwen2.5-VL-7B) and Med-LVLMs (e.g., Lingshu-7B), which are trained on more medical data with diverse modalities. 
Furthermore, TumorChain shows better pxunrformance than commercial models (Claude3-Haiku, Gemini2.0-Flash).
Nevertheless, the $CoT_e$ score is slightly lower than the performance of the latest GPT5-mini, demonstrating its strong capabilities in the medical domain.

% \begin{wrapfigure}[13]{t}{0.5\textwidth}
%   \centering
%     \includegraphics[width=0.5\textwidth]{figs/leida.pdf}
%   \caption{(a) Effect of Classification Loss Weight. (b) Performance Comparison Across Five Organs.}
%   \label{fig:leida}
% \end{wrapfigure}

\subsection{Ablation and Generalization Studies}
\textbf{Effect of CoT and IIR.} To comprehensively evaluate the impact of our proposed CoT reasoning and Interleaved Iterative Reasoning (IIR) approaches, we conducted systematic ablation studies. As shown in Table \ref{tab:tumor_bench}, lines 13 to 16, the combined use of CoT and IIR yields substantial performance gains across all subtasks, with an overall accuracy improvement of 5.64\% compared to the baseline. This highlights that focusing on ROI region features markedly enhances model performance in 3D image analysis. Models employing CoT alone consistently improved results on most tasks, notably advancing clinical interpretability without sacrificing analytical accuracy. We further assessed inference latency after incorporating the IIR mechanism. The segmentation model incurs only a minimal increase in inference time (2.51 seconds per sample), while delivering an approximately 4\% accuracy boost. The significant benefit of emphasizing ROI features far outweighs the slight time cost, strongly validating the reliability and effectiveness of our method for intelligent tumor analysis. 

\begin{figure*}[t]
  \centering
  \includegraphics[width=1\linewidth]{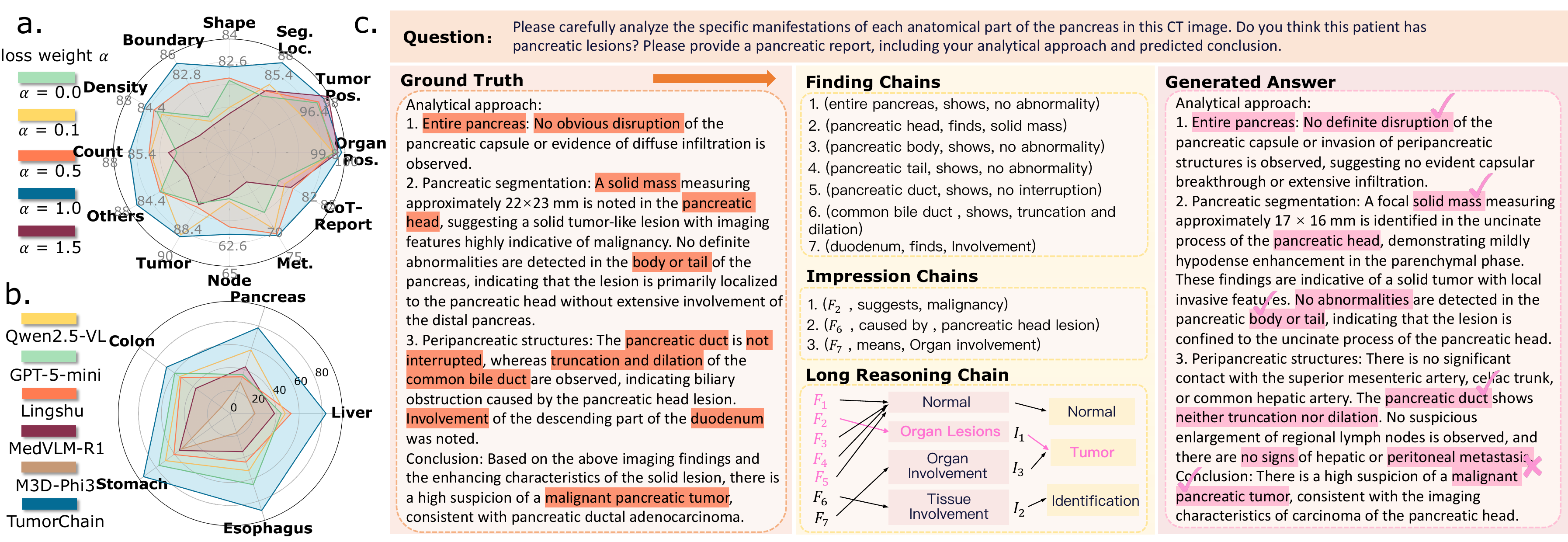}
    % \vspace{mm}
    \caption{(a) Ablation on classification loss weight $\alpha$. (b) Model performance on five organs of \texttt{TumorCoT}. (c) Workflow of reasoning chains extraction and evaluation for task 4.}
    \vspace{-4mm}
  \label{fig:case}
\end{figure*}

\textbf{Impact of Classification Loss Weight.}
We systematically investigated the model’s performance under different classification loss weight settings, as illustrated in Figure \ref{fig:case} (a). The results indicate that appropriately increasing the classification loss weight (e.g., $\alpha$=1.0) leads to significant performance improvements across multiple tasks, achieving the highest average accuracy of 84.41\%. However, setting the weight too low or too high (such as $\alpha$=0.0 or $\alpha$=1.5) results in diminished performance, further demonstrating the essential role of loss weight adjustment in the collaborative optimization of hybrid models. Detailed results are available in Appendix~\ref{sec:appendix-experiment_Result_details}.

\begin{wrapfigure}[13]{t}{0.44\textwidth}
  \centering
      \vspace{-4mm}
    \includegraphics[width=0.44\textwidth]{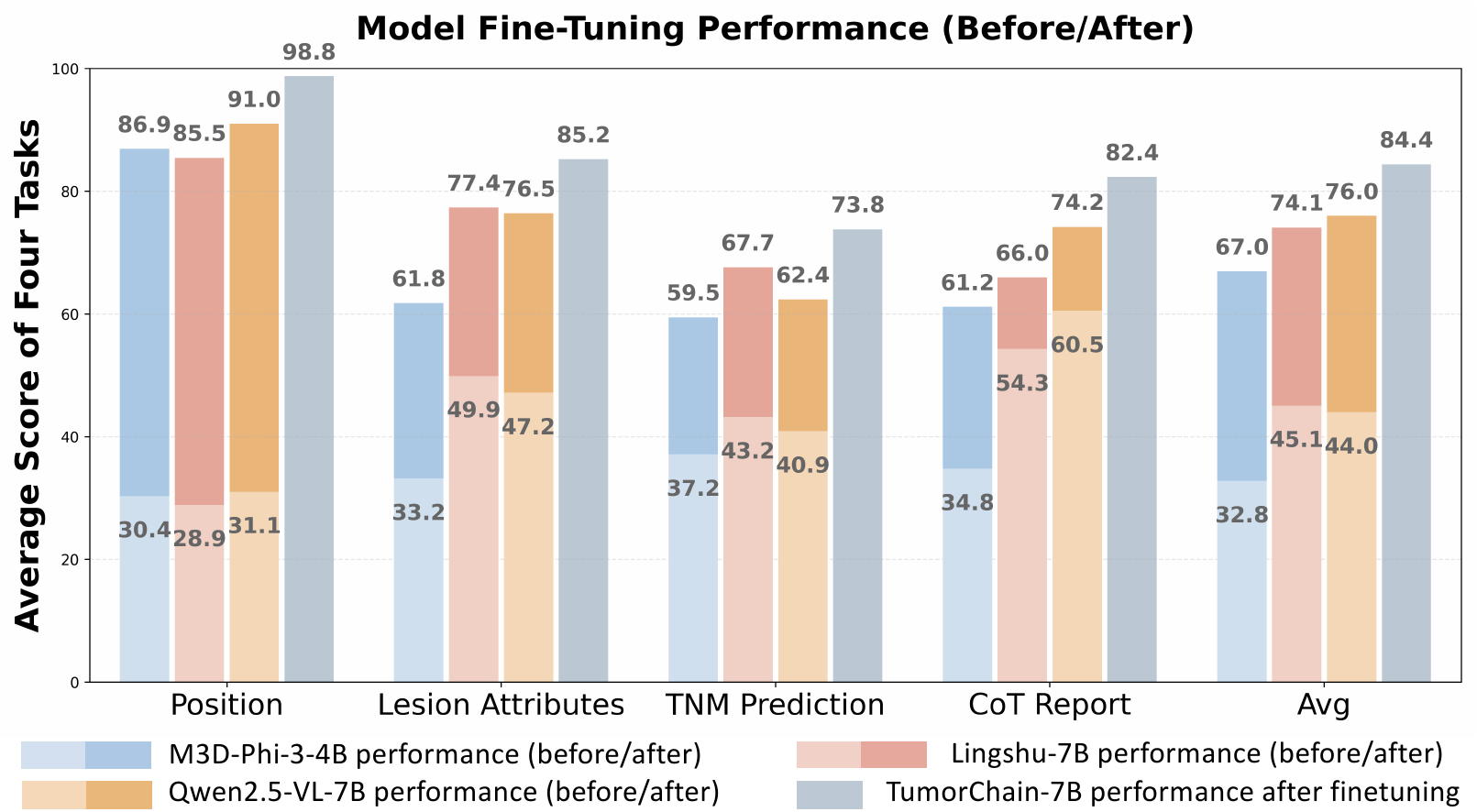}
    \vspace{-2mm}
  \caption{Baseline model improvements by finetuning with TumorCoT-1.5M.}
  \label{fig:finetune}
\end{wrapfigure}

\textbf{Effect of finetuning by TumorCoT-1.5M train set.} To more directly demonstrate the independent contribution of our dataset to model finetuning, we selected three representative baseline models for finetuning on our TumorCoT-1.5M training set: the 3D medical LVLM M3D-Phi-3-4B, the 2D medical LVLM Lingshu-7B, and the 2D general-purpose model Qwen2.5-VL-7B. Figure~\ref{fig:finetune} shows the performance changes of these baseline models before and after finetuning (see Appendix~\ref{baseline-finetune} for detailed results). The results indicate that all baseline models achieve significant improvements after finetuning, demonstrating the unique and important contribution of our dataset in the tumor domain. Meanwhile, our proposed TumorChain-7B still substantially surpasses the finetuned baselines, further highlighting the effectiveness of our model design.

\begin{wraptable}[14]{r}{0.44\textwidth} 
    \centering
    \vspace{-5mm}
    \caption{Results on DeepTumorVQA.}
    \resizebox{0.43\textwidth}{!}{% 
    \begin{tabular}{lcccc}
        \toprule
        \textbf{Model} & \textbf{Recog.} & \textbf{V.R.} & \textbf{M.R.} & \textbf{Avg.} \\
        \midrule
        \rowcolor{gray!5}
        \multicolumn{5}{c}{\textit{Commercial Models}} \\
        \midrule
        Claude3-Haiku      & 38.70  & 31.53  & 34.81  & 35.01 \\
        Gemini2.0-Flash    & 50.88  & 33.81  & 35.06  & 39.92 \\
        GPT-5-mini         & 49.29  & 35.57  & 29.87  & 38.24 \\
        \midrule
        \rowcolor{gray!5}
        \multicolumn{5}{c}{\textit{Open-Source Models}} \\
        \midrule
        Qwen2.5-VL-7B     & 45.31 & 36.94 & 37.04 & 39.76 \\
        InternVL-2.5-8B   & 50.06 & 31.35 & \underline{44.20} & 41.87 \\
        MiniCPM-V4.5-9B   & 50.14 & 33.69 & 42.86 & 42.23 \\
        \midrule
        HealthGPT-3.8B    & 43.32 & 30.68 & 39.75 & 37.92 \\
        Lingshu-7B        & 50.24 & 36.19 & 37.78 & 41.40 \\
        MedVLM-R1-2B      & \underline{56.41} & 38.27 & 33.33 & \underline{42.67} \\
        \midrule
        RadFM             & 50.19  & \underline{39.82}  & 30.87  & 40.29 \\
        M3D-Phi-3-4B       & 29.53  & 23.22  & 23.70  & 25.48 \\
        \rowcolor{blue!5}
        \textbf{TumorChain-7B}    & \textbf{73.30} & \textbf{53.31} & \textbf{45.93} & \textbf{57.51} \\
        % \rowcolor{blue!5}
        % + \textbf{DeepTumorVQA}     &   &   &   &   \\
        % \rowcolor{blue!5}
        % + \textbf{DeepTumorVQACoT}  &   &   &   &   \\
        \bottomrule
    \end{tabular}
    }  
    \label{tab:deeptumorVQA}
\end{wraptable}

% \subsection{Generalization on DeepTumorVQA Benchmark.}
\textbf{Generalization on DeepTumorVQA Benchmark.}
% 为评估我们提出的模型和方法在肿瘤分析领域的泛化能力，我们在最新的公开肿瘤基准数据集 DeepTumorVQA 上进行了全面的实验。如表4所示，TumorChain-7B在病灶识别任务达到了73.30的准确率，在视觉和医学推理上面的性能相比主流商业模型和开源模型均显著领先，同时平均准确率相比排名第二的推理模型MedVLM-R1高了14.84%。这些结果有力证明了我们方法在开放域肿瘤分析任务中的卓越泛化能力和稳健性。
To evaluate the generalization capability of our model and methods in tumor analysis, 
we conducted comprehensive experiments on the latest public tumor benchmark, DeepTumorVQA. As shown in Table \ref{tab:deeptumorVQA}, without seeing any data from this dataset, TumorChain-7B achieved an accuracy of 73.30\% in lesion recognition and demonstrated substantially better performance in both visual and medical reasoning compared to leading commercial and open-source models. In addition, its average accuracy exceeds that of the second-best reasoning model, MedVLM-R1, by 14.84\%. These results strongly demonstrate the exceptional generalization ability and robustness of our approach in open-domain tumor analysis tasks.

\section{related work}
\noindent\textbf{Medical Large Vision-Language Models.} Med-LVLMs~\cite{li2025eyecaregpt,xie2025heartcare} such as LLaVA-Med~\citep{li2023llava}, Lingshu~\citep{xu2025lingshu} and HealthGPT~\citep{lin2025healthgpt} have rapidly advanced multimodal clinical decision support by aligning ViT-based 2D image features with language models. Methods like M3D~\citep{bai2024m3d}, CT-Chat~\citep{hamamci2024foundation} and Merlin~\citep{blankemeier2024merlin} incorporate 3D CNNs and ViTs to better capture volumetric data, enabling more precise analysis of clinical 3D images. However, these Med-LVLMs lack causal reasoning for complex cases, prompting a shift toward CoT and knowledge-guided reasoning paradigms. MedVLThinker~\citep{huang2025medvlthinker} and MedVLM-R1~\citep{pan2025medvlm} combine CoT reasoning with reinforcement learning to improve interpretability, while ReasonMed~\citep{sun2025reasonmed} optimizes reasoning paths via multi-agent verification. MedResearcher-R1~\citep{yu2025medreseacher} enhances deep reasoning by leveraging knowledge graphs for multi-hop QA sample construction. By emphasizing logical chains, medical vision reasoning models are overcoming traditional 2D and 3D VLM limitations and opening new directions for explainable and reliable clinical decision-making.
We further summarize related work with our TumorChain in three aspects: \textbf{Medical Benchmark}, \textbf{Medical Large Vision-Language Models}, and \textbf{Online Interleaved Reasoning}. The detailed discussions are shown in Appendix~\ref{sec:appendix-related_work}.

\section{Conclusion}
We introduce \textbf{TumorChain}, a multimodal framework designed for comprehensive clinical tumor reasoning, spanning radiological findings, study-level impressions, and pathology predictions. 
% By leveraging hybrid-model interleaved reasoning, TumorChain aligns global and local features with organ-level clinical priors, reducing hallucination risk and improving interpretability in medical inference. 
We also construct \texttt{TumorCoT-1.5M}, the largest multimodal dataset for tumor reasoning, and propose a benchmark protocol to evaluate stepwise reasoning fidelity. 
Extensive experiments demonstrate TumorChain’s superior performance across diverse tumor-related tasks, benefiting the advancement of clinical AI in oncology. Furthermore, Appendix~\ref{sec:appendix-cases} presents case studies and qualitative error analyses as useful references for future model improvement and clinical practice. 

\section*{Acknowledgments}
This work has been supported in part by the NSFC (No. 62436007), the China Postdoctoral Science Foundation under Grant Number 2024M752794, the ZJNSF (No. LZ25F020004), the Key Research and Development Projects in Zhejiang Province (No. 2025C01128, 2025C01030, 2025C02156),
Ningbo Yongjiang Talent Introduction Programme (2023A400-G).

% \newpage
\section*{Ethics statement}
This study used hospital-held clinical imaging data and was reviewed and approved by the hospital Ethics Committee. Data collection and use complied with applicable laws, regulations, and ethical standards. All data were de-identified and anonymized prior to use, with personally identifiable information removed. Access to the data was restricted to authorized personnel, and appropriate safeguards (e.g., encryption and access controls) were implemented to prevent disclosure of personal information. The data were used solely for the purposes of this research.

\section*{Reproducibility Statement}
To advance safe, explainable, and reproducible multimodal reasoning for high-stakes tumor analysis, detailed information about our project can be found on our project homepage at \\ \href{https://github.com/ZJU4HealthCare/TumorChain}{https://github.com/ZJU4HealthCare/TumorChain}. Detailed descriptions of hyperparameters and experimental settings can be found in Appendix~\ref{sec:appendix-training_details}.

\bibliographystyle{assets/plainnat}
\bibliography{paper}

% \clearpage
\newpage
\beginappendix
\section*{Appendix}
\label{sec:appendix}
\Cref{sec:appendix-LLM}. LLM Usage Statement.

\Cref{sec:appendix-notation}. Notation Table.

\Cref{sec:appendix-related_work}. Related Work.

\Cref{sec:appendix-implementation}. Experiments and Implementation Details.

\quad \Cref{sec:appendix-training_details}. Training Details.

\quad \Cref{sec:appendix-accuracy}. Semantic Accuracy Metric by GPT5.

\quad \Cref{sec:appendix-experiment_Result_details}. Experiment Result Details.

\Cref{sec:appendix-data_engine}. Interactive-Validated COT Data Engine.

\quad \Cref{sec:appendix-agents}. Data Engine Components Details.

\quad \Cref{sec:appendix-kg}. Diagnostic Knowledge Graph.

% \quad \Cref{sec:appendix-template}. The Template of Structured Reports.

\quad \Cref{sec:appendix-prompt}. Prompt Design of Data Engine.

\quad \Cref{sec:appendix-organ}. Multi-Organ Masks and Merged Organ Masks.

\quad \Cref{sec:Formal-evaluation}. Formal, quantitative evaluation of the CoT data engine by expert clinicians.

\Cref{sec:appendix-cot_eval}. CoT Evaluation Details: TumorChain-Eval.

\quad \Cref{sec:appendix-extraction}. Stage 1: Extraction of Medical Entity–Relation-Entity Triples.

\quad \Cref{sec:appendix-protocol}. Stage 2: CoT Evaluation Protocol.

\Cref{sec:appendix-cases}. Cases Study.

\newpage

\section{LLM Usage Statement}
\label{sec:appendix-LLM}
In this work, we employ LLMs in three strictly controlled ways. First, LLMs are utilized to rephrase the source data during dataset construction in the CoT data engine~\ref{sec:appendix-data_engine}), , thereby constructing a high-quality training dataset based on the original information while preserving the clinical context. Second, we leverage LLMs as experts to evaluate our model against baselines (see Appendix~\ref{sec:appendix-cot_eval}), following clearly defined criteria for consistency and fairness. Third, we employ LLMs to identify and correct grammatical errors in our manuscript, ensuring clarity and linguistic accuracy throughout the paper.

\section{Notation Table}
\label{sec:appendix-notation}
To facilitate understanding and ensure consistency of symbols used throughout the paper of TumorChain , Table \ref{tab:notation} provides a concise summary of all key notations, offering readers a quick reference for the variables and operators involved.

\begin{table}[ht]
   \small
    \centering
    \renewcommand{\arraystretch}{1.2}
    \caption{Notation in the TumorChain Pipeline.}
    \label{tab:notation}
    \resizebox{1\columnwidth}{!}{
        \begin{tabular}{c|l}
    \toprule
 Notation & Description\\
  \midrule
$\mathcal{F}(\cdot)$  & Represents the TumorChain framework.\\
$\mathcal{{E}}_{v}(\cdot)$  & 3D vision encoder\\
$\mathcal{S}eg(\cdot)$ &  Organ segmentation expert  \\
$\mathcal{C}ls(\cdot)$  &  Auxiliary classification model \\
$\mathcal{P}(\cdot)$ &  Multi-layer perceptron (MLP) projector\\
$\mathcal{LLM}(\cdot)$ & Backbone of LLM \\
$\mathcal{V}_{ct} \in \mathbb{R}^{H\times W\times D}$ & Represents a 3D CT volume of height $H$, width $W$, and depth $D$.  \\
$\mathcal{T}_{task}$ & Input of task prompt text \\
$\mathcal{R}_{cot}$ & The output of TumorChain \\
$\tau_v$ & A series of vision tokens after 3D encoder $\mathcal{E}_v$\\
$\tau_{g}$  & Global CT volume tokens (projector-aligned into LLM space). \\
$\mathcal{M}_{organ}$ & Denotes the organ mask generated by the segmentation model   $\mathcal{S}eg(\cdot)$  \\
$\tau_{l}$  & Local vision tokens \\
$\Lambda(\cdot)$ &  Denotes the organ matching operation \\
$\Gamma(\cdot)$ & Represents the operation of extracting local token $\tau_{l}$ of size $L_l\times K$ according to mask.\\
${y}$ &  Result after Classification layer $Cls(\tau_l)$\\
$ \tau_{in} = [\tau_g, \mathcal{T}_{task}, \mathcal{T}^1, \tau_{l}^1, \mathcal{T}^i, \tau_l^i,...]$ &  Indicates all input tokens of LLM \\
$\mathcal{T}_{task}$   & Task text prompt tokens \\
$\tau_l^i$ & Task target organ tokens \\
$\mathcal{T}^i$. $i\in [1, N]$  &  Represents the N potential related or ROI organs/suborgans reasoned by LLM.  \\
$\mathcal{R}^1_{cot}$ & The initial diagnostic result output by the LLM\\
$M_l^1$ &  Local ROI mask \\
$L_{total}$ &  Training loss \\
$N$ & Represents the output text length \\
$M$ & represents the sample number \\
$\alpha$ & Loss weight \\
    \bottomrule
        \end{tabular}
        }
\end{table}

\section{related work}
\label{sec:appendix-related_work}
% \subsection{Medical Benchmark}
\noindent
\textbf{Medical Benchmark.}
% With the growing demand for multimodal reasoning in clinical applications, 
Early public medical benchmarks such as VQA-RAD~\citep{lau2018dataset}, VQA-Med~\citep{ben2019vqa}, SLAKE~\citep{liu2021slake}, and PathVQA~\citep{he2020pathvqa} have greatly advanced LVLM development in healthcare. These datasets are primarily 2D and feature simple, template-based QA pairs. Recent works employ data synthesis to build larger and more multimodal benchmarks: RadFM~\citep{wu2023generalistfoundationmodelradiology} generates expert-verified questions from literature and cases. OmniMedVQA~\citep{hu2024omnimedvqa} leverages category-driven templates with GPT-4~\citep{achiam2023gpt} for paraphrasing and distractor creation. HealthBench~\citep{arora2025healthbench} combines synthetic generation and adversarial testing for multi-turn, multilingual dialogues. In the field of tumor analysis, DeepTumorVQA~\citep{chen2025vision} focuses on fine-grained 3D CT tumor detection. However, most benchmarks remain limited to multiple-choice formats and basic reasoning, insufficient to meet clinical demands for interpretability and in-depth analysis.

% \subsection{Medical Large Vision-Language Models}
\noindent
\textbf{Medical Large Vision-Language Models.}
Med-LVLMs such as LLaVA-Med~\citep{li2023llava}, Lingshu~\citep{xu2025lingshu}, Med-PaLM~\citep{singhal2025toward} and HealthGPT~\citep{lin2025healthgpt} have rapidly advanced multimodal clinical decision support by aligning ViT-based 2D image features with language models. Methods like M3D~\citep{bai2024m3d}, CT-Chat~\citep{hamamci2024foundation} and Merlin~\citep{blankemeier2024merlin} incorporate 3D CNNs and ViTs to better capture volumetric data, enabling more precise analysis of clinical 3D images. However, these Med-LVLMs lack causal reasoning for complex cases, prompting a shift toward CoT and knowledge-guided reasoning paradigms~\cite{zhang2022boostmis}. MedVLThinker~\citep{huang2025medvlthinker} and MedVLM-R1~\citep{pan2025medvlm} combine CoT reasoning with reinforcement learning to improve interpretability, while ReasonMed~\citep{sun2025reasonmed} optimizes reasoning paths via multi-agent verification. MedResearcher-R1~\citep{yu2025medreseacher} enhances deep reasoning by leveraging knowledge graphs for multi-hop QA sample construction. By emphasizing logical chains, medical vision reasoning models are overcoming traditional 2D and 3D VLM limitations and opening new directions for explainable and reliable clinical decision-making.

% \subsection{Online interleaved reasoning }
\noindent
\textbf{Online Interleaved Reasoning.}
Online interleaved reasoning (IR) dynamically alternates thinking and answering, enabling more efficient multi-hop reasoning than traditional methods~\citep{xie2025interleaved,trivedi2022interleaving}. IRCoT~\citep{trivedi2022interleaving} shows that interleaving retrieval, reasoning, and response generation significantly improves performance on complex tasks. Chain-of-Focus~\citep{zhang2025chain} further introduces adaptive visual search and zooming with reinforcement learning to enhance multimodal reasoning. CX-Mind~\citep{li2025cx} adopts interleaved reasoning strategies for chest X-ray diagnosis, generating verifiable reasoning chains. Online IR aligns Med-LVLM decision-making with clinical workflows, motivating us to explore IR for 3D tumor analysis.

\section{Experiments and Implementation Details}
\label{sec:appendix-implementation}

\subsection{Training Details}
\label{sec:appendix-training_details}
\textbf{Data Preparation and Preprocessing.} TumorChain takes as input complete three-dimensional CT scans in \texttt{.nii.gz} format. It is important to note that these volumetric images exhibit considerable variability in spatial dimensions along the x, y, and z axes. To ensure consistency and model stability, we employ a multi-step preprocessing pipeline.

Initially, we extract the soft tissue window level and window width for each CT scan according to standard radiological protocols, followed by normalization of voxel intensities. This step reduces inter-scan variability and facilitates robust feature extraction. Given the heterogeneity in image shapes, we address dimensional inconsistency by applying precise cropping and zero-padding strategies. These operations are carefully performed to prevent geometric distortion and loss of anatomical information.Finally, all processed CT volumes are resized to a fixed shape of (256, 256, 32)(height × width × depth), in order to meet the input requirements of the 3D vision encoder. This standardized preprocessing framework ensures that all input volumes possess the same spatial resolution, thereby improving model compatibility and computational efficiency.

\textbf{Model Architecture.} 
%例如学习率，优化器，batch size， scheduler等等
We adopt Qwen2.5-VL-3B and Qwen2.5-VL-7B as the base multimodal backbones for TumorChain-3B and TumorChain-7B, respectively. Since the original Qwen-VL vision encoder only supports 2D images, we replace it with M3D encoder~\citep{bai2024m3d} to support volumetric 3D CT inputs. To bridge visual and language modalities, we employ a two-layer multilayer perceptron (MLP) projection module, consisting of two Linear layers with an intermediate ReLU activation, to map the extracted organ-level visual embeddings into the LLM token space and facilitate multimodal fusion. For the large language model component, the pretrained LLM backbone of Qwen2.5-VL-3B/7B is used for high-level multimodal reasoning and report generation. Additionally, an auxiliary classification head is implemented as a single fully connected layer to perform binary classification (normal vs. abnormal) on each segmented organ region, providing supervisory calibration. Both TumorChain models are trained under identical hyperparameter configurations, including learning rate and precision settings, following the hybrid model fine-tuning strategy described in the main text.

\begin{table}[b]
   \small
   \vspace{-1mm}
    \centering
    \renewcommand\tabcolsep{20pt}
    \caption{Hyperparameter Configuration}
    \label{tab:hyperparameter}
    \resizebox{0.5\columnwidth}{!}{
        \begin{tabular}{c|c}
    \toprule
 Hyperparameter & Value \\
  \midrule
Optimizer& AdamW \\
Learning rate& $3 \times 10^{-5}$ \\
Learning rate scheduler& cosine decay \\
Weight decay& 0.0 \\
Warmup ratio& 0.03 \\
Mixed precision& bf16 (fp16 disabled) \\
Gradient accumulation steps& 2 \\
Per-device batch size& 2 \\
Number of dataloader workers& 16 \\
Number of epochs& 1.0 \\
% Gradient checkpointing& enabled \\
    \bottomrule
        \end{tabular}
        }
\end{table}

\textbf{Hardware and Distributed Training.}
Model training is conducted on 32 NVIDIA A800 GPUs. For the 3B model, training is performed for 12 hours, whereas the 7B model is trained for 16 hours. The training set comprises 1.5M CT samples, utilizing 90\% of the available data. All models are trained using the DeepSpeed distributed framework (configuration: ZeRO-2), effectively enabling large-scale parallel training and efficient memory utilization.

\textbf{Hyperparameter Configuration.} All training and evaluation pipelines are conducted with strict reproducibility control. Detailed hyperparameter settings are as shown in table~\ref{tab:hyperparameter}.

% \textbf{Hyperparameter Configuration.} Detailed hyperparameter settings are as follows:
% \begin{itemize}
%     \item \textbf{Backbone LLM}: Qwen2.5-7B-Instruct
%     \item \textbf{Vision Encoder}: M3D-CLIP
%     \item \textbf{Per-device batch size}: 2
%     \item \textbf{Gradient accumulation steps}: 2
%     \item \textbf{Mixed precision}: BF16 (fp16 disabled)
%     \item \textbf{Number of epochs}: 1.0
%     \item \textbf{Learning rate}: $3 \times 10^{-5}$
%     \item \textbf{Weight decay}: 0.0
%     \item \textbf{Optimizer}: AdamW
%     \item \textbf{Learning rate scheduler}: Cosine decay
%     \item \textbf{Warmup ratio}: 0.03
%     \item \textbf{Gradient checkpointing}: Enabled
%     \item \textbf{Number of dataloader workers}: 16
%     \item \textbf{Logging interval}: Every 5 steps
%     \item \textbf{Evaluation}: Disabled during training 
%     \item \textbf{Training stage}: Supervised fine-tuning (SFT)
%     \item \textbf{Experiment tracking}: SwanLab
% \end{itemize}

\subsection{Semantic Accuracy Metric by GPT5}
\label{sec:appendix-accuracy}
To assess the quality of open-ended answers in the VQA task for CT imaging, we introduce a semantic consistency metric specifically designed for shape, boundary, density, count, TNM stage prediction, and other clinical questions. For each question–answer pair, the evaluator first classifies the question type, identifies its key clinical focus, and then compares the model’s answer to the ground truth (reference answer). The metric judges the response as correct if the essential medical meaning and main clinical finding are preserved, regardless of minor differences in expression or reasoning process. This approach enables reliable evaluation of the model’s clinical interpretability and factual accuracy, focusing on whether the core diagnostic content is communicated consistently. The semantic consistency metric ensures that the assessment is robust against linguistic variation and prioritizes agreement in clinical understanding over superficial textual similarity.

\begin{figure}
\centering
\includegraphics[width=1\textwidth]{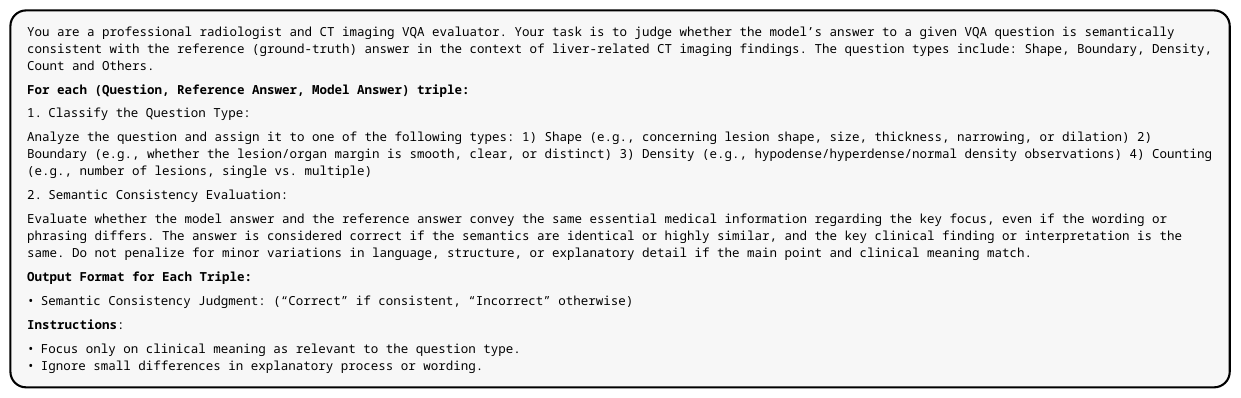}
\vspace{-2mm}
\caption{Semantic Consistency Evaluation Prompt.}
%% kp: can mention our specific technique in the caption, e.g., H$^2$LoRA, describe a bit, etc.
\vspace{-3mm}
\label{fig:metric_gpt5}
\end{figure}

An illustrative example of the evaluation prompt used to guide semantic consistency assessment is shown in Figure~\ref{fig:metric_gpt5}.
% 这里放正文里面放不下gpt5的表格

\subsection{Experiment Result Details}
\label{sec:appendix-experiment_Result_details}
\subsubsection{Experimental results by organ category}
In this section, we present more detailed ablation studies and comprehensive Chain-of-Thought (CoT) evaluation results than those included in the main text. Table~\ref{tab:cot_eval_split} and Table~\ref{tab:deeptumorVQA_appendix} summarize the organ-level accuracy of our model and various baselines on the TumorCoT and DeepTumorVQA benchmarks, respectively. The scope of our evaluation focuses on the five major digestive organs: liver, stomach, pancreas, colon, and esophagus. For generalization tests, we specifically analyze the overlapping three organs—liver, pancreas, and colon—within the DeepTumorVQA dataset, as these closely match our benchmark task settings.

\begin{table}[t]
\caption{CoT-Report comparison of TumorChain with other \textbf{Reasoning VLMs} on \texttt{TumorCoT}.}
\label{tab:cot_eval_split}
\begin{center}
 \renewcommand\tabcolsep{18pt}
\resizebox{\textwidth}{!}{
\begin{tabular}{lcccccc}
\toprule
\textbf{Reasoning Models} &
\textbf{Liver} & \textbf{Pancreas} & \textbf{Colon} &
\textbf{Stomach} & \textbf{Esophagus} & \textbf{Average} \\
\midrule
Claude3-Haiku   & 34.70 & 56.13 & 44.08 & 17.11 & 39.77 & 38.36 \\
Gemini2.0-Flash  & 42.35 & 29.03 & 56.24 & 71.15 & 38.64 & 47.48 \\
GPT-5-Mini       & 44.48 & 35.71 & 59.76 & 76.85 & 65.12 & 56.38 \\
\midrule
Qwen2.5-VL-7B    & 47.94 & 58.15 & 54.60 & 69.79 & 52.41 & 56.58 \\
InternVL-2.5-8B  & 48.89 & 49.20 & 48.89 & 55.84 & 52.54 & 51.07 \\
MiniCPM-V-4.5-9B & 46.07 & 43.29 & 45.69 & 50.26 & 47.59 & 46.58 \\
\midrule
HealthGPT-3.8B   & 48.69 & 63.74 & 36.44 & 31.74 & 45.81 & 45.28 \\
Lingshu-7B       & 53.03 & 33.55 & 53.23 & 60.71 & 44.42 & 48.99 \\
MedVLM-R1-2B     & 38.58 & 42.78 & 37.18 & 54.79 & 34.90 & 41.65 \\
\midrule
RadFM            & 11.34 & 15.42 &  2.50 & 35.20 & 13.71 & 15.63 \\
M3D-Phi3-4B      & 21.51 & 28.76 &  9.51 & 52.57 & 17.72 & 26.01 \\
\midrule
\rowcolor{blue!5} 
\textbf{TumorChain-3B} & \underline{69.89} & \underline{72.20} & \underline{67.39} & \underline{91.63} & \underline{64.34} & \underline{73.09} \\
\rowcolor{blue!5} 
\textbf{TumorChain-7B} & \textbf{83.45} & \textbf{78.43} & \textbf{68.65} & \textbf{93.45} & \textbf{88.71} & \textbf{82.54} \\
\bottomrule 
\end{tabular}
}
\end{center}
\end{table}
\begin{table}[t]
\caption{Comparison of TumorChain with other VLMs on the public DeepTumorVQA benchmark.}
\label{tab:deeptumorVQA_appendix}
\begin{center}
\resizebox{0.714\textwidth}{!}{
\renewcommand\tabcolsep{18pt}
\renewcommand{\arraystretch}{1.0}
\begin{tabular}{lcccc}
\toprule
\textbf{Model} & \textbf{Liver} & \textbf{Pancreas} & \textbf{Colon} & \textbf{Average}\\
\midrule
\rowcolor{gray!5}
\multicolumn{5}{c}{\textit{Generalist Models}} \\
\midrule
Qwen2.5-VL-7B     & 36.32 & 32.39 & 46.33 & 38.35 \\
InternVL-2.5-8B   & 34.52 & 35.86 & 52.75 & 41.04 \\
MiniCPM-V4.5-9B   & 40.02 & \underline{45.20} & 52.93 & 46.05 \\
\midrule
\rowcolor{gray!5}
\multicolumn{5}{c}{\textit{Commercial Models}} \\
\midrule
Claude3-Haiku     & 32.67 & 32.62 & 18.40 & 27.90 \\
Gemini2.0-Flash   & 35.33 & 35.02 & 44.80 & 38.38 \\
GPT-5-mini        & \underline{41.20} & 42.48 & \underline{73.60} & \underline{52.43} \\
\midrule
\rowcolor{gray!5}
\multicolumn{5}{c}{\textit{2D Medical Models}} \\
\midrule
HealthGPT-3.8B    & 30.32 & 30.26 & 53.56 & 38.05 \\
Lingshu-7B        & 38.39 & 38.01 & 50.68 & 42.36 \\
MedVLM-R1-2B      & 35.98 & 36.73 & 69.85 & 47.52 \\
\midrule
\rowcolor{gray!5}
\multicolumn{5}{c}{\textit{3D Medical Models}} \\
\midrule
RadFM             & 31.97 & 34.11 & 55.71 & 40.60 \\
M3D-Phi3-4B       & 24.82 & 24.88 & 19.68 & 23.13 \\
\rowcolor{blue!5}
\textbf{TumorChain-7B} & \textbf{55.41} & \textbf{62.45} & \textbf{84.33} & \textbf{67.40} \\
% \rowcolor{blue!5}
% + \textbf{DeepTumorVQA}     &   &   &   &   &   \\
% \rowcolor{blue!5}
% + \textbf{DeepTumorVQACoT}  &   &   &   &   &   \\
\bottomrule
\end{tabular}
}
\end{center}
\end{table}

\subsubsection{Ablation experiments under different classification loss weight}

\begin{table*}[t]
\centering
\caption{Ablation study of the effect of classfication loss weight $\alpha$ in TumorChain.}
\label{tab:clsloss_ablation}
\resizebox{\textwidth}{!}{
\begin{tabular}{cccccccccccccc}
\toprule
\multirow{3}{*}{\textbf{$\alpha$}} 
& \multicolumn{2}{c}{\textbf{Position}} 
& \multicolumn{6}{c}{\textbf{Lesion Attributes}} 
& \multicolumn{3}{c}{\textbf{TNM Prediction}} 
& \multirow{3}{*}{\textbf{\begin{tabular}{@{}c@{}}CoT-\\Report\end{tabular}}}
& \multirow{3}{*}{\textbf{Avg.}} \\
\cmidrule(rl){2-3} \cmidrule(rl){4-9} \cmidrule(rl){10-12} 
& \begin{tabular}{@{}c@{}}Organ\\Pos.\end{tabular} 
& \begin{tabular}{@{}c@{}}Tumor\\Pos.\end{tabular} 
& \begin{tabular}{@{}c@{}}Seg.\\Loc.\end{tabular} 
& Shape & Boundary & Density & Count & Others 
& Tumor & Node & Met. &  &  \\
\midrule
\textbf{0.0} & 99.92 & 97.01 & 82.46 & 81.45 & 75.82 & 83.01 & 80.89 & 82.35 & 86.11 & 57.88 & 65.22 & 77.41 & 80.79 \\
\textbf{0.1} & 99.90 & 95.58 & 84.01 & 79.78 & 75.09 & 82.10 & 83.85 & 81.85 & 88.69 & 58.16 & 69.54 & 78.29 & 81.40 \\
\textbf{0.5} & 99.90 & 97.14 & 83.14 & 81.60 & 81.15 & 83.51 & 84.01 & 82.45 & 86.43 & 60.85 & 70.65 & 79.54 & 82.53 \\
\rowcolor{blue!5} \textbf{1.0} & 99.97 & 97.57 & 86.88 & 82.28 & 84.52 & 85.05 & 86.20 & 86.57 & 88.83 & 61.63 & 71.07 & 82.36 & 84.41 \\
\textbf{1.5} & 99.94 & 97.91 & 83.21 & 79.40 & 74.46 & 75.37 & 81.83 & 77.41 & 86.23 & 57.51 & 57.46 & 79.21 & 79.16 \\
\bottomrule
\end{tabular}}
\end{table*}

Table~\ref{tab:clsloss_ablation} reports the results of ablation experiments under different classification loss weight ($\alpha$) settings. We systematically investigate the impact of varying $\alpha$, and the findings show that moderately increasing the classification loss weight (e.g., $\alpha$=1.0) yields significant performance improvements across multiple tasks, with a highest average accuracy of 84.41\%. In contrast, setting $\alpha$ either too low or too high leads to reduced performance. These results highlight the critical role of appropriate loss weight adjustment in optimizing the collaborative training of hybrid models.

\subsubsection{Ablation experiments on baselines before and after fine-tuning}
\begin{table*}[t]
\centering
\caption{Ablation study of baselines before and after finetuning on TumorCoT.}
\label{tab:finetune_ablation}
\resizebox{\textwidth}{!}{
\begin{tabular}{lccccc}
\toprule
\multicolumn{6}{c}{\textbf{Avg. Score (Before finetuning\,/\,After finetuning)}} \\
\midrule
\textbf{Model} & \textbf{Position} & \textbf{Lesion Attributes} & \textbf{TNM Prediction} & \textbf{CoT Report} & \textbf{Avg.} \\
\midrule
M3D-Phi-3-4B      & 30.36\,/\,\textbf{86.92} & 33.23\,/\,\textbf{61.78} & 37.16\,/\,\textbf{59.48} & 34.79\,/\,\textbf{61.18} & 32.84\,/\,\textbf{66.98} \\
Lingshu-7B        & 28.93\,/\,\textbf{85.46} & 49.91\,/\,\textbf{77.36} & 43.21\,/\,\textbf{67.65} & 54.30\,/\,\textbf{65.98} & 45.10\,/\,\textbf{74.11} \\
Qwen2.5-VL-7B     & 31.05\,/\,\textbf{91.02} & 47.17\,/\,\textbf{76.45} & 40.94\,/\,\textbf{62.37} & 60.54\,/\,\textbf{74.21} & 44.04\,/\,\textbf{76.01} \\
\rowcolor{blue!5} TumorChain-7B  & -\,/\,\textbf{98.77}   & -\,/\,\textbf{85.25}   & -\,/\,\textbf{73.84}   & -\,/\,\textbf{82.36}   & -\,/\,\textbf{84.41} \\
\bottomrule
\end{tabular}}
\end{table*}

\label{baseline-finetune}
Our contributions include both a large-scale, high-quality dataset and a novel model architecture. Accordingly, our experiments cover three aspects: main evaluations demonstrating overall performance, ablation studies assessing the impact of individual model components, and generalization tests exploring robustness across different data distributions. The main results in the paper demonstrate significant improvements across multiple metrics. In addition, we conducted rigorous ablation studies on three key components of the architecture—Impression-Information Retrieval (IIR), classification-loss weighting, and CoT-formatted data. The results indicate that each module delivers clinically meaningful performance gains.

To further isolate and validate the contribution of the dataset itself, we fine-tuned three representative baseline models on our training set to facilitate direct performance comparisons and thoroughly assess the gains attributable to the dataset alone. The supplemental results, shown as Table~\ref{tab:finetune_ablation}, reveal that all baseline models achieve notable improvements after fine-tuning, underscoring the unique and substantial value of our dataset within the tumor domain. 

Meanwhile, our proposed TumorChain-7B continues to significantly outperform these fine-tuned baselines, reflecting the effectiveness of our architectural innovations.

\section{Interactive-Validated COT Data Engine}
\label{sec:appendix-data_engine}
\subsection{Data Engine Components Details}
\label{sec:appendix-agents}
\textbf{(i) Diagnostic Knowledge Graph.} In collaboration with five organ-specialist radiologists, we construct a triplet-based ("entity–relation–entity") knowledge graph from diagnostic guidelines, textbooks, and representative cases, covering anatomy, findings, impressions, histopathology, and risk factors (see Appendix Figure~\ref{fig:kg}). All organ segmentation standards follow international conventions, enabling hierarchical substructure and tumor-grade analysis. During reasoning chain construction, the data agent retrieves relevant nodes and relations from the knowledge graph to ensure traceable and reliable logic and minimize factual and logical errors.
\begin{figure}
\centering
\includegraphics[width=1\textwidth]{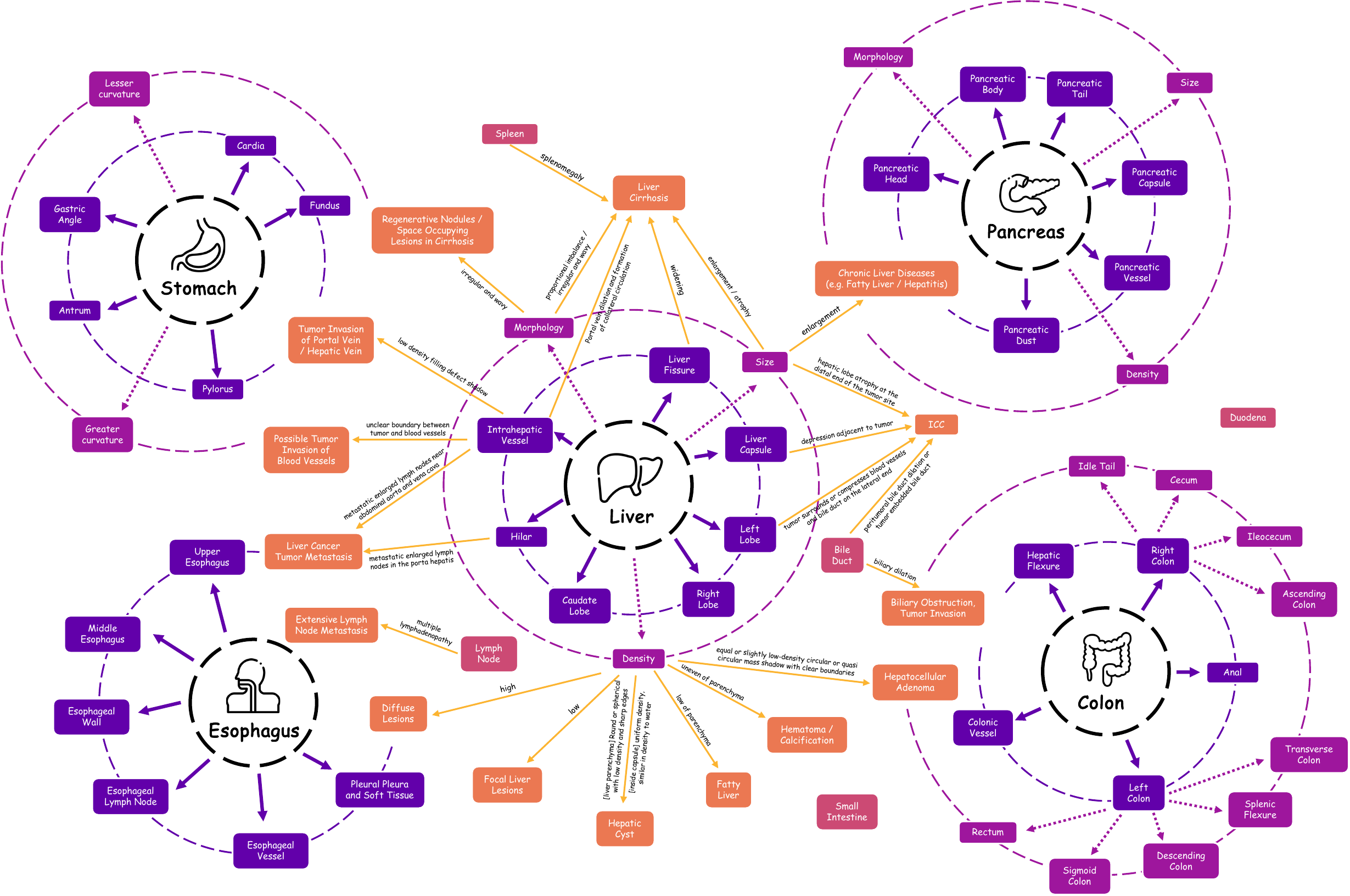}
\vspace{-2mm}
\caption{The overview of the diagnostic knowledge graph of 5 organs by doctors.}
%% kp: can mention our specific technique in the caption, e.g., H$^2$LoRA, describe a bit, etc.
\vspace{-3mm}
\label{fig:kg}
\end{figure}

\textbf{(ii) Structured Feature Extractor.} We employ Qwen3-235B-A22B as the structured feature extractor to perform rigorous cleaning and terminology standardization across multi-source, multi-format data, with all imaging descriptors aligned to the RadLex international standard. Leveraging organ segmentation from the knowledge graph, the module automatically extracts substructure-level descriptions and structured features—including clinical information (age, gender), lesion attributes (location, shape, margin, density, count), and six additional categories—from CT and pathology reports. TNM staging information and final diagnostic conclusions are also extracted from pathology reports, ensuring the completeness and accuracy of subsequent reasoning chains.

\textbf{(iii) Segmentation Expert Model.} We utilize TotalSegmentator to segment 117 organs and merge them to 56 organs (see Appendix Table \ref{tab:seg_labels}). The organ and tumor masks of five digestive organs are refined by radiologists, enabling structured ROI localization for subsequent VQA tasks.

\textbf{(iv) Traceable CoT Reasoner.} 
GPT-4o-mini, selected for strong language and medical expertise, receives multidimensional structured reports and knowledge graph links, generating organ- and lesion-level VQA that conforms to medical guidelines.

\textbf{(v) CoT Logic Calibrator.} During reasoning chain generation, Claude3.5-Haiku automatically verifies logical integrity. Upon detecting inconsistencies, the system uses two prompt strategies: expanding organ regions via segmentation for re-reasoning, or requesting further clarification and self-reflection from the reasoning module.

\textbf{(vi) Summarizer for Finding-Impression-Pathology Chains.} We employ GPT-5-mini to aggregate all reasoning chains. This module receives structured information from pathology reports to construct a higher-level pathology analysis VQA. If inconsistencies arise between reasoning chains and pathological conclusions, the system automatically reinitiates inference. Upon successful validation, it extracts findings and impressions related to staging, lymph node involvement, and metastasis, generating TNM prediction VQA pairs. Finally, a traceable CoT-format report is produced for each organ, facilitating standardized and interpretable reasoning outputs.

\subsection{Diagnostic Knowledge Graph}
\label{sec:appendix-kg}
As figure~\ref{fig:kg} shows, the diagnostic knowledge graph constructed in this study serves as the core clinical prior for the TumorCoT-1.5M dataset and TumorChain model, providing structured, traceable medical knowledge support for multimodal CoT reasoning in tumor analysis. Unlike general medical knowledge graphs that cover broad healthcare domains, this KG is tumor-centric and organ-specific, focusing exclusively on the five major digestive organs (esophagus, stomach, colon, pancreas, and liver) closely associated with clinical tumor diagnosis and treatment. Its core value lies in bridging the gap between unstructured medical text (e.g., radiology reports) and structured reasoning logic, ensuring that every step of the CoT reasoning process (from imaging findings to pathology predictions) is grounded in evidence-based medical knowledge rather than arbitrary inference.

The construction of the KG adheres to a multi-source, hierarchical verification principle, integrating three levels of authoritative data sources to balance comprehensiveness, accuracy, and clinical relevance. \textbf{A-level} primarily includes international and domestic clinical diagnostic guidelines of five organs, which define the core knowledge boundaries of the KG, including TNM staging criteria and standard imaging features of malignant tumors. \textbf{B-level} covers classic textbooks and high-impact academic papers. These sources supplement critical anatomical details and advanced imaging features. \textbf{C-level} consists of real-world expert-annotated cases, including typical tumor cases labeled by 5 experienced organ-specialist radiologists and teaching cases from the Department of Radiology at top-tier tertiary hospitals.

This design not only ensures that the knowledge conforms to international clinical guidelines but also incorporates real-world diagnostic experience from expert physicians, making it more suitable for high-stakes tumor analysis scenarios.
% 来源：A级别：五大器官影像诊断指南
% B级别：学术论文、教材资料（医学影像诊断学第五版）
% C级别: 医生标注的典型案例，教学案例
% 构造方式：与五名五大器官影像诊断的专业医生合作共同标注
% 包含实体类型：器官，器官子结构分段，病灶属性，影像结论，疾病类型等等

% \subsection{The Template of Structured Reports}
% \label{sec:appendix-template}

\subsection{Prompt Design of Data Engine}
\label{sec:appendix-prompt}

% 这里放画好的prompt示例，简单介绍一下task2的prompt design，参考附录c1部分的reasoner组件的细节

When designing prompt templates for each agent in our data engine, we adhere to the following principles to ensure both data quality and clinical validity.

\textbf{Standardization of Medical Terminology.} All descriptions are required to follow internationally recognized medical terminology standards (e.g., RadLex Radiological Lexicon, AJCC TNM staging system). This prevents ambiguity and ensures consistency across tasks (for example, using “hypodense lesion” instead of the vague expression “low-density shadow”).

\textbf{Mandatory Reasoning Chain.} Each response must consist of two components: a “Reasoning Process” and a “Summary.” The reasoning process should explicitly reflect clinical diagnostic logic, typically progressing from organ localization to feature observation and finally to pathological correlation. The summary should present the diagnostic conclusion in a standardized form, e.g., “Thus, the answer is ...”.

\textbf{Task Boundary Specification.} To avoid cross-task redundancy, prompts are designed to clearly delineate the input scope for different subtasks (e.g., lesion localization tasks focus exclusively on anatomical positioning without involving density assessment or malignancy determination).

\textbf{Multi-Source Information Integration.} Prompts are required to guide the agent to incorporate both structured medical knowledge (e.g., anatomical ontologies, tumor metastasis pathways) and quantitative image-derived features (e.g., lesion boundary, density values). This ensures that the reasoning process is grounded in objective evidence rather than subjective speculation.

\begin{figure}
\centering
\includegraphics[width=1\textwidth]{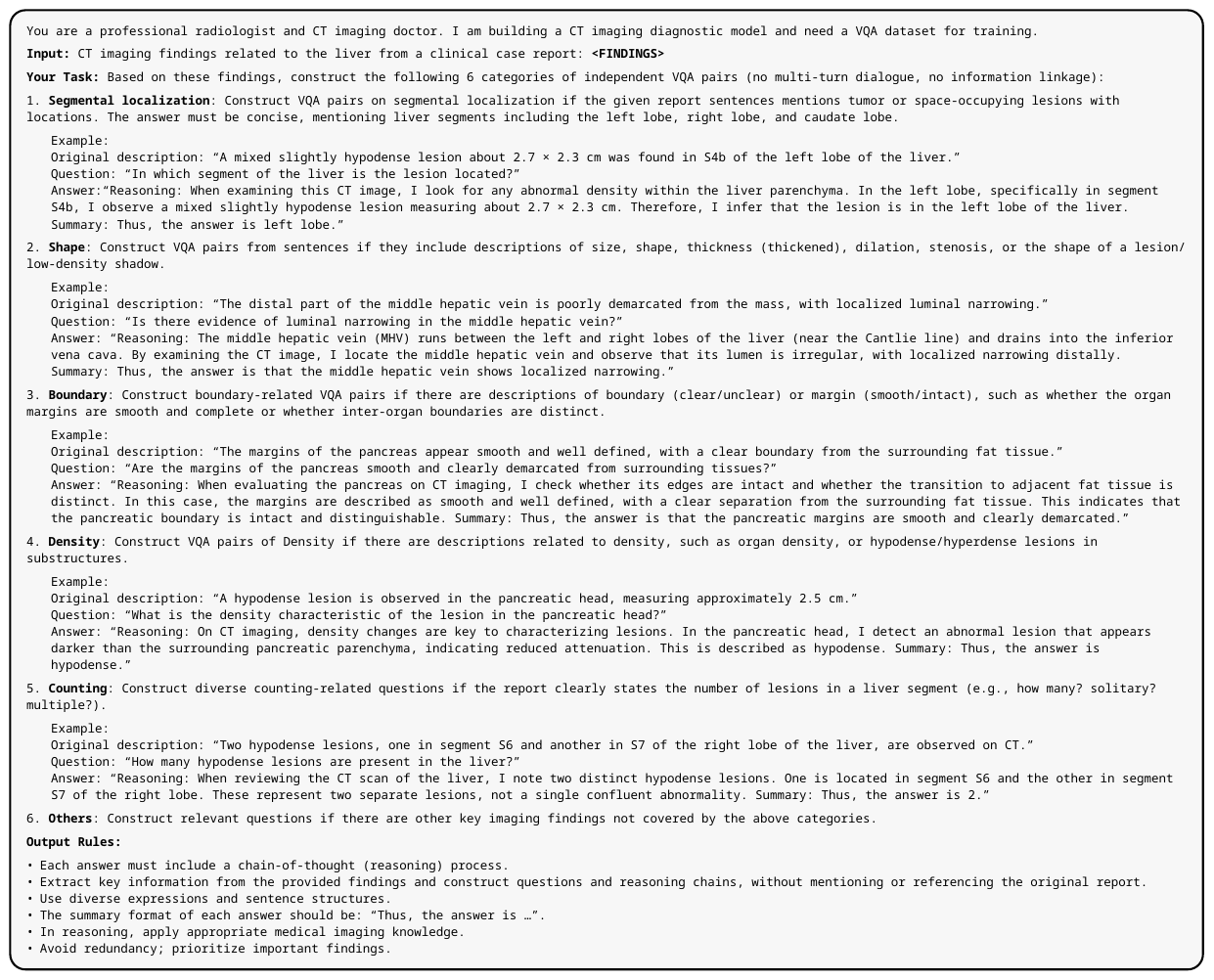}
\vspace{-2mm}
\caption{Prompt example: Traceable CoT Reasoner for extracting lesion attributes.}
%% kp: can mention our specific technique in the caption, e.g., H$^2$LoRA, describe a bit, etc.
\vspace{-3mm}
\label{fig:prompt}
\end{figure}

Following these principles, the CoT inference framework aligns with clinical diagnostic standards and effectively supports the four core tasks of tumor analysis: anatomical localization, lesion attribute characterization, TNM staging prediction, and structured report generation. An illustrative example of a prompt designed for the Traceable CoT Reasoner to extract lesion attributes is shown in Figure~\ref{fig:prompt}.

\subsection{Multi-Organ Masks and Merged Organ Masks}
\label{sec:appendix-organ}
Table~\ref{tab:seg_labels} provides the detailed organ mask IDs from TotalSegmentor~\citep{wasserthal2023totalsegmentator} and the mapping with our merged organ mask IDs.
\begin{scriptsize}
\renewcommand{\arraystretch}{1.1}
\begin{longtable}{c|c|c|c}
\caption{Mapping Table of TotalSegmentator Organ Indexes After Aggregation.}
\label{tab:seg_labels}\\
\toprule
\rowcolor{gray!5}
\textbf{New Label} & \textbf{Segmentator Name} & \textbf{Included Organs} & \textbf{TotalSeg. Label} \\
\hline
0 & background & background & 0 \\
\hline
1 & spleen & spleen & 1 \\
\hline
2 & kidney\_right & kidney\_right & 2 \\
\hline
3 & kidney\_left & kidney\_left & 3 \\
\hline
4 & gallbladder & gallbladder & 4 \\
\hline
5 & liver & liver & 5 \\
\hline
6 & stomach & stomach & 6 \\
\hline
7 & pancreas & pancreas & 7 \\
\hline
8 & adrenal\_gland\_right & adrenal\_gland\_right & 8 \\
\hline
9 & adrenal\_gland\_left & adrenal\_gland\_left & 9 \\
\hline
10 & lung\_upper\_lobe\_left & lung\_upper\_lobe\_left & 10 \\
\hline
11 & lung\_lower\_lobe\_left & lung\_lower\_lobe\_left & 11 \\
\hline
12 & lung\_upper\_lobe\_right & lung\_upper\_lobe\_right & 12 \\
\hline
13 & lung\_middle\_lobe\_right & lung\_middle\_lobe\_right & 13 \\
\hline
14 & lung\_lower\_lobe\_right & lung\_lower\_lobe\_right & 14 \\
\hline
15 & esophagus & esophagus & 15 \\
\hline
16 & trachea & trachea & 16 \\
\hline
17 & thyroid\_gland & thyroid\_gland & 17 \\
\hline
18 & small\_bowel & small\_bowel & 18 \\
\hline
19 & duodenum & duodenum & 19 \\
\hline
20 & colorectum & colon & 20 \\
\hline
21 & urinary\_bladder & urinary bladder & 21 \\
\hline
22 & prostate & prostate & 22 \\
\hline
\multirow{2}{*}{23} & \multirow{2}{*}{kidney\_cyst} & kidney\_cyst\_left & 23 \\
                     &                               & kidney\_cyst\_right & 24 \\
\hline
24 & vertebrae\_S1 & vertebrae\_S1 & 26 \\
\hline
\multirow{5}{*}{25} & \multirow{5}{*}{lumbar\_vertebrae} & vertebrae\_L5 & 27 \\
                     &                                   & vertebrae\_L4 & 28 \\
                     &                                   & vertebrae\_L3 & 29 \\
                     &                                   & vertebrae\_L2 & 30 \\
                     &                                   & vertebrae\_L1 & 31 \\
\hline
\multirow{12}{*}{26} & \multirow{12}{*}{thoracic\_vertebrae} & vertebrae\_T12 & 32 \\
                     &                                      & vertebrae\_T11 & 33 \\
                     &                                      & vertebrae\_T10 & 34 \\
                     &                                      & vertebrae\_T9  & 35 \\
                     &                                      & vertebrae\_T8  & 36 \\
                     &                                      & vertebrae\_T7  & 37 \\
                     &                                      & vertebrae\_T6  & 38 \\
                     &                                      & vertebrae\_T5  & 39 \\
                     &                                      & vertebrae\_T4  & 40 \\
                     &                                      & vertebrae\_T3  & 41 \\
                     &                                      & vertebrae\_T2  & 42 \\
                     &                                      & vertebrae\_T1  & 43 \\
\hline
\multirow{7}{*}{27} & \multirow{7}{*}{cervical\_vertebrae} & vertebrae\_C7 & 44 \\
                     &                                     & vertebrae\_C6 & 45 \\
                     &                                     & vertebrae\_C5 & 46 \\
                     &                                     & vertebrae\_C4 & 47 \\
                     &                                     & vertebrae\_C3 & 48 \\
                     &                                     & vertebrae\_C2 & 49 \\
                     &                                     & vertebrae\_C1 & 50 \\
\hline
28 & sacrum & sacrum & 25 \\
\hline
\multirow{2}{*}{29} & \multirow{2}{*}{humerus} & humerus\_left & 69 \\
                     &                         & humerus\_right & 70 \\
\hline
\multirow{2}{*}{30} & \multirow{2}{*}{scapula} & scapula\_left & 71 \\
                     &                         & scapula\_right & 72 \\
\hline
\multirow{2}{*}{31} & \multirow{2}{*}{clavicula} & clavicula\_left & 73 \\
                     &                           & clavicula\_right & 74 \\
\hline
\multirow{2}{*}{32} & \multirow{2}{*}{femur} & femur\_left & 75 \\
                     &                       & femur\_right & 76 \\
\hline
\multirow{2}{*}{33} & \multirow{2}{*}{hip} & hip\_left & 77 \\
                     &                     & hip\_right & 78 \\
\hline
\multirow{12}{*}{34} & \multirow{12}{*}{rib\_left} & rib\_left\_1 & 92 \\
                     &                            & rib\_left\_2 & 93 \\
                     &                            & rib\_left\_3 & 94 \\
                     &                            & rib\_left\_4 & 95 \\
                     &                            & rib\_left\_5 & 96 \\
                     &                            & rib\_left\_6 & 97 \\
                     &                            & rib\_left\_7 & 98 \\
                     &                            & rib\_left\_8 & 99 \\
                     &                            & rib\_left\_9 & 100 \\
                     &                            & rib\_left\_10 & 101 \\
                     &                            & rib\_left\_11 & 102 \\
                     &                            & rib\_left\_12 & 103 \\
\hline
\multirow{12}{*}{35} & \multirow{12}{*}{rib\_right} & rib\_right\_1 & 104 \\
                      &                             & rib\_right\_2 & 105 \\
                      &                             & rib\_right\_3 & 106 \\
                      &                             & rib\_right\_4 & 107 \\
                      &                             & rib\_right\_5 & 108 \\
                      &                             & rib\_right\_6 & 109 \\
                      &                             & rib\_right\_7 & 110 \\
                      &                             & rib\_right\_8 & 111 \\
                      &                             & rib\_right\_9 & 112 \\
                      &                             & rib\_right\_10 & 113 \\
                      &                             & rib\_right\_11 & 114 \\
                      &                             & rib\_right\_12 & 115 \\
\hline
36 & sternum & sternum & 116 \\
\hline
37 & costal\_cartilages & costal\_cartilages & 117 \\
\hline
38 & heart & heart & 51 \\
\hline
39 & aorta & aorta & 52 \\
\hline
40 & pulmonary\_vein & pulmonary\_vein & 53 \\
\hline
41 & brachiocephalic\_trunk & brachiocephalic\_trunk & 54 \\
\hline
\multirow{2}{*}{42} & \multirow{2}{*}{subclavian\_artery} & subclavian\_artery\_right & 55 \\
                      &                                   & subclavian\_artery\_left & 56 \\
\hline
\multirow{2}{*}{43} & \multirow{2}{*}{common\_carotid\_artery} & common\_carotid\_artery\_right & 57 \\
                      &                                       & common\_carotid\_artery\_left  & 58 \\
\hline
\multirow{2}{*}{44} & \multirow{2}{*}{brachiocephalic\_vein} & brachiocephalic\_vein\_left & 59 \\
                      &                                      & brachiocephalic\_vein\_right & 60 \\
\hline
45 & atrial\_appendage\_left & atrial\_appendage\_left & 61 \\
\hline
46 & superior\_vena\_cava & superior\_vena\_cava & 62 \\
\hline
47 & inferior\_vena\_cava & inferior\_vena\_cava & 63 \\
\hline
48 & portal\_vein\_and\_splenic\_vein & portal\_vein\_and\_splenic\_vein & 64 \\
\hline
\multirow{2}{*}{49} & \multirow{2}{*}{iliac\_artery} & iliac\_artery\_left & 65 \\
                     &                               & iliac\_artery\_right & 66 \\
\hline
\multirow{2}{*}{50} & \multirow{2}{*}{iliac\_vena} & iliac\_vena\_left & 67 \\
                     &                             & iliac\_vena\_right & 68 \\
\hline
\multirow{6}{*}{51} & \multirow{6}{*}{gluteus} & gluteus\_maximus\_left & 80 \\
                     &                          & gluteus\_maximus\_right & 81 \\
                     &                          & gluteus\_medius\_left & 82 \\
                     &                          & gluteus\_medius\_right & 83 \\
                     &                          & gluteus\_minimus\_left & 84 \\
                     &                          & gluteus\_minimus\_right & 85 \\
\hline
\multirow{2}{*}{52} & \multirow{2}{*}{autochthon} & autochthon\_left & 86 \\
                     &                            & autochthon\_right & 87 \\
\hline
\multirow{2}{*}{53} & \multirow{2}{*}{iliopsoas} & iliopsoas\_left & 88 \\
                     &                           & iliopsoas\_right & 89 \\
\hline
54 & spinal\_cord & spinal\_cord & 79 \\
\hline
55 & brain & brain & 90 \\
\hline
56 & skull & skull & 91 \\
\bottomrule
\end{longtable}
\end{scriptsize}

\subsection{Quantitative evaluation of the CoT data engine by expert clinicians.}
\label{sec:Formal-evaluation}
To systematically assess the factual correctness and clinical plausibility of the generated reasoning chains, we randomly sampled 5,000 VQA instances from the TumorChain-1.5M dataset across task types and submitted them to a team led by a board-certified radiologist(8 years of oncologic imaging experience) for evaluation from two perspectives:

$\bullet$ \textbf{Usability}: The clinician first assessed whether the VQA result matches the corresponding information in the original medical report. If the VQA result is correct and can provide reliable support in clinical practice, it is marked as usable; otherwise, it is marked as unusable.

$\bullet$ \textbf{Clinical Reasonableness}: For all usable data, we further categorized the samples:

a. High quality: For usable cases, the reasoning chain must exhibit completeness, scientific soundness, and logical coherence; only when all criteria are satisfied is it marked as high quality.

b. Acceptable: For usable cases where the overall logic is sound but some steps are slightly brief or marginally insufficient, it is marked as acceptable.

Evaluation results show that the usability rate reaches 95.88\%, and among all usable samples, 97.85\% are high-quality. These results validate the effectiveness of our data-generation pipeline and demonstrate the reliability and practical clinical value of the dataset.

\section{CoT Evaluation Details: TumorChain-Eval}
\label{sec:appendix-cot_eval}
This appendix provides a comprehensive description of the evaluation framework for Chain-of-Thought (CoT) reasoning. The framework of \textbf{TumorChain-Eval} is organized into two stages. In the first stage, structured (subject, relation, object) triples are extracted from medical texts to serve as factual building blocks for reasoning. In the second stage, the generated CoT reasoning chains are assessed for quality based on these triples.
\subsection{Stage 1: Extraction of Medical Entity–Relation-Entity Triples}
\label{sec:appendix-extraction}
This stage serves as a prerequisite for CoT evaluation. Its goal is to convert unstructured medical text into structured knowledge representations, thereby supplying accurate and unambiguous factual units for subsequent reasoning-chain assessment.
\paragraph{Task Definition.}
Given a medical radiology report, the objective is to extract all (subject, relation, object) triples that capture medical facts.
\textit{Subject}: Typically an organ (e.g., liver), an anatomical structure (e.g., pericolonic vessels), or a lesion (e.g., enlarged lymph node).
\textit{Relation}: A verb or state term that describes the medical semantic relationship between the subject and the object (e.g., not observed, suggests, absent, supports).
\textit{Object}: Generally a medical finding, lesion (e.g., wall thickening, abnormal density), condition (e.g., mass-like change), or another related organ.

\begin{table}[htbp]
\centering
\caption{Prompts for medical triples extraction in stage 1 of CoT evaluation}
\label{tab:medical_prompt}

\begin{tabular}{p{0.95\linewidth}}
\toprule
\textbf{Prompt for Medical Fact Triple Extraction} \\
\midrule

You are a medical fact--extraction assistant specialized in parsing
knowledge from the given clinical text.
Your task is to analyze the input and extract all factual triples
$(\textit{subject}, \textit{relation}, \textit{object})$,
where each triple expresses a single clear medical fact.
Ensure that all information is presented strictly in triple form,
focusing on \textbf{organs/structures}, \textbf{relations}, and
\textbf{lesions}.

\textbf{Input Example:}

First, examine the right hemicolon, cecum, ascending colon, and hepatic
flexure; no wall thickening or abnormal density is observed, indicating
no definite mass or inflammatory change in the right hemicolon.

Second, check the left hemicolon, transverse colon, and splenic flexure,
descending colon, sigmoid colon, and rectum; again no abnormal
thickening or abnormal density is found, supporting the absence of
significant lesions in the left hemicolon.

Third, evaluate colon-related internal structures and surrounding
tissues; no pericolonic vascular abnormality is detected, and the liver,
gallbladder, pancreas, spleen, kidneys, and pelvic structures show no
colon-related significant lesion, enlarged lymph nodes, or free fluid.

These findings collectively do not support tumor or active inflammatory
changes. Overall, the imaging features do not suggest colorectal cancer
or other qualifying pathology.

\textbf{Output Requirements:}

In each triple, the first and third elements must be an organ or lesion,
and the second element must describe their relationship.
If the second and third positions are reversed, correct them.

The output format should be:

$(\text{Right hemicolon}, \text{not observed}, \text{wall thickening});$

\\
\bottomrule
\end{tabular}

\end{table}

\paragraph{Prompt Design and Specification}
To guide the model in performing triple extraction, we design standardized prompts tailored to the characteristics of medical tasks. The prompt clearly specifies the model’s role, the extraction objective, and the required output format, while providing illustrative examples to facilitate structured responses.
The prompt for LLM to extract medical entity-relation-entity triples is shown in Table \ref{tab:medical_prompt}.

\subsection{Stage 2: CoT Evaluation Protocol}
Following the triple-extraction stage, this phase aims to comprehensively assess the quality of the model-generated Chain-of-Thought (CoT) reasoning chains. Evaluation is performed by comparing model predictions (Pred) against the reference ground truth (GT).
\label{sec:appendix-protocol}

\begin{center}

\renewcommand{\arraystretch}{1.25} % 行间距

\begin{table}[]
\caption{Finding Chain (FC) scoring criteria.}
\centering
\small
% 列规范：注意没有在最前面和最后面加 '|'，因此无最左/最右竖线
\resizebox{\textwidth}{!}{
\begin{tabular}{m{1cm}|m{3cm}|m{4.4cm}|m{7cm}}
\hline
\textbf{Chain Level} & \textbf{Scoring Dimension} & \textbf{Description} & \textbf{Scoring Criteria} \\
\hline
\multirow{18}{*}{\centering \rotatebox{90}{Finding Chain}} 
  & \multirow{6}{*}{Existence Match} 
  & \multirow{6}{*}{\parbox{4.5cm}{\raggedright Degree to which predicted facts match the ground-truth (GT) facts.}} 
  & \textbf{10}: Prediction contains all key GT facts with no omission or redundancy.\\ \cline{4-4}
  &&& \textbf{8--9}: Prediction covers the vast majority of GT facts, with only minor omissions or redundancies (<10\%).\\ \cline{4-4}
  &&& \textbf{6--7}: Prediction covers some GT facts but exhibits moderate omissions or redundancies (10--30\%).\\ \cline{4-4}
  &&& \textbf{4--5}: Prediction covers only a small portion of GT facts; substantial omissions or redundancies (30--50\%).\\ \cline{4-4}
  &&& \textbf{1--3}: Low match rate; only a few facts are correct and the error rate is very high.\\ \cline{4-4}
  &&& \textbf{0}: Prediction includes none of the GT facts.\\ \cline{2-4}

  & \multirow{6}{*}{Completeness}
  & \multirow{6}{*}{\parbox{4.5cm}{\raggedright Extent to which key facts are fully expressed without missing or spurious content.}}
  & \textbf{10}: Prediction covers 100\% of GT facts with no omissions.\\ \cline{4-4}
  &&& \textbf{8--9}: Only minor omissions ($<$10\%), high coverage.\\ \cline{4-4}
  &&& \textbf{6--7}: Noticeable omissions (10--30\%), coverage is moderately compromised.\\ \cline{4-4}
  &&& \textbf{4--5}: Majority of key facts are missing (30--50\%), poor coverage.\\ \cline{4-4}
  &&& \textbf{1--3}: Extensive omissions ($>$50\%), very low coverage.\\ \cline{4-4}
  &&& \textbf{0}: Prediction fails to cover any key facts.\\ \cline{2-4}

  & \multirow{6}{*}{Accuracy}
  & \multirow{6}{*}{\parbox{4.5cm}{\raggedright Correctness of factual statements in the prediction.}}
  & \textbf{10}: All predicted entries are accurate with no invalid or incorrect statements.\\ \cline{4-4}
  &&& \textbf{8--9}: Vast majority accurate ($<$10\% problematic), only minor issues.\\ \cline{4-4}
  &&& \textbf{6--7}: Moderate errors or invalid entries (10--30\%) that negatively affect overall quality.\\ \cline{4-4}
  &&& \textbf{4--5}: Numerous erroneous entries ($>$30\%), only a small fraction correct.\\ \cline{4-4}
  &&& \textbf{1--3}: Very few accurate entries ($>$50\% error rate).\\ \cline{4-4}
  &&& \textbf{0}: All predictions are invalid; no correct entries.\\
\hline
\end{tabular}}

\label{tab:s1_scoring}
\end{table}

\end{center}

\paragraph{Reasoning-Chain Categorization.}
Both the predicted (Pred) and reference (GT) reasoning chains are segmented into three hierarchical levels to enable fine-grained evaluation:
\textit{(i) Finding Chain (FC):} Consists of basic facts directly extracted from radiological descriptions (e.g., no wall thickening observed, soft-tissue mass detected). These represent independent, objective observations without inferential steps.
\textit{(ii) Impression Chain (IC):}  Comprises intermediate medical impressions or preliminary suggestions derived from multiple S1 findings (e.g., suggests localized inflammatory change). This level reflects simple, initial reasoning.
\textit{(iii) Long Reasoning Chain (LRC):} Represents high-level medical reasoning and conclusions that integrate all findings (S1) and impressions (S2) (e.g., consistent with imaging features of malignant tumor). This chain must exhibit complete logical derivation and clinical diagnostic value.

\paragraph{Scoring Criteria.}
We employ GPT-4o to score the FC, IC, and LRC sub-chains along multiple dimensions. All metrics are rated on a 10-point scale.

The scoring criteria of FC are shown in Table \ref{tab:s1_scoring}.

The scoring criteria of IC are shown in Table \ref{tab:s2_scoring}.

The scoring criteria of LRC are shown in Table \ref{tab:s3_scoring}.

% \vspace{1em}

\begin{center}

\renewcommand{\arraystretch}{1.25} % 行间距
\begin{table}[]
\centering
\small
\caption{Impression Chain (IC) scoring criteria.}
% 列规范：无最左和最右竖线
\resizebox{\textwidth}{!}{
\begin{tabular}{m{1cm}|m{3cm}|m{4.4cm}|m{7cm}}
\hline
\textbf{Chain Level} & \textbf{Scoring Dimension} & \textbf{Description} & \textbf{Scoring Criteria} \\
\hline
\multirow{18}{*}{\centering \rotatebox{90}{Impression Chain}}
  & \multirow{6}{*}{Clarity}
  & \multirow{6}{*}{\parbox{4.5cm}{\raggedright Medical clarity of the impression statement.}}
  & \textbf{10}: Impression is completely clear, logically coherent, and unambiguous.\\ \cline{4-4}
  &&& \textbf{8--9}: Description is clear with only minor incomplete expressions or slight ambiguity ($<$10\%).\\ \cline{4-4}
  &&& \textbf{6--7}: Basically clear but contains some ambiguities that moderately hinder understanding.\\ \cline{4-4}
  &&& \textbf{4--5}: Blurry description with numerous ambiguities significantly affecting medical interpretation.\\ \cline{4-4}
  &&& \textbf{1--3}: Very difficult to understand; almost unusable due to severe ambiguity.\\ \cline{4-4}
  &&& \textbf{0}: Impression chain is empty or entirely invalid and unclear.\\ \cline{2-4}

  & \multirow{6}{*}{Consistency}
  & \multirow{6}{*}{\parbox{4.5cm}{\raggedright Logical consistency of the impression with the underlying FINDING chain.}}
  & \textbf{10}: Fully consistent with the factual chain; all impressions are derived from facts with no unreasonable content.\\ \cline{4-4}
  &&& \textbf{8--9}: Overall consistent with only minor ($<$10\%) deviations from the factual chain.\\ \cline{4-4}
  &&& \textbf{6--7}: Partially inconsistent with the factual chain, showing moderate deviation (10--30\%).\\ \cline{4-4}
  &&& \textbf{4--5}: Large proportion of content inconsistent with or weakly related to the factual chain (30--50\%).\\ \cline{4-4}
  &&& \textbf{1--3}: Impression content is mostly illogical and unrelated to the factual chain.\\ \cline{4-4}
  &&& \textbf{0}: Impression is entirely invalid or completely contradicts the factual chain.\\ \cline{2-4}

  & \multirow{6}{*}{Medical Utility}
  & \multirow{6}{*}{\parbox{4.5cm}{\raggedright Clinical usefulness of the impression for diagnosis and decision-making.}}
  & \textbf{10}: Impression chain is highly useful, directly supporting diagnosis and clinical decision-making with no additional input needed.\\ \cline{4-4}
  &&& \textbf{8--9}: High clinical utility; most content is medically meaningful with only minor adjustments required.\\ \cline{4-4}
  &&& \textbf{6--7}: Partial diagnostic value but considerable portions lack utility or have vague meaning (10--30\%).\\ \cline{4-4}
  &&& \textbf{4--5}: Very limited clinical utility ($>$50\% of content lacks diagnostic value).\\ \cline{4-4}
  &&& \textbf{1--3}: Impression provides almost no diagnostic significance or is largely incorrect.\\ \cline{4-4}
  &&& \textbf{0}: Impression chain is invalid or contains no medically meaningful statements.\\
\hline
\end{tabular}}

\label{tab:s2_scoring}
\end{table}

\end{center}
% \vspace{1em}

\begin{center}
\renewcommand{\arraystretch}{1.25} % 行间距
\begin{table}[]
\centering
\small
\caption{Long Reasoning Chain (LRC) scoring criteria.}
% 列规范：无最左和最右竖线
\resizebox{\textwidth}{!}{
\begin{tabular}{m{1cm}|m{3cm}|m{4.4cm}|m{7cm}}
\hline
\textbf{Chain Level} & \textbf{Scoring Dimension} & \textbf{Description} & \textbf{Scoring Criteria} \\
\hline
\multirow{24}{*}{\centering \rotatebox{90}{Long Reasoning Chain}}
  & \multirow{6}{*}{Logical Completeness}
  & \multirow{6}{*}{\parbox{4.5cm}{\raggedright Logical closure and completeness of higher-order reasoning.}}
  & \textbf{10}: The reasoning chain perfectly covers all key points, with no logical gaps or omissions.\\ \cline{4-4}
  &&& \textbf{8--9}: Reasoning is largely complete, with only minor ($<$10\%) logical gaps or omitted details.\\ \cline{4-4}
  &&& \textbf{6--7}: Some logical interruptions or omissions exist, but most reasoning paths remain valid (10--30\%).\\ \cline{4-4}
  &&& \textbf{4--5}: Significant logical gaps, missing many key points, and notable interruptions in reasoning (30--50\%).\\ \cline{4-4}
  &&& \textbf{1--3}: Most of the reasoning chain is invalid; significant logical flaws, key derivations incomplete.\\ \cline{4-4}
  &&& \textbf{0}: No reasoning process or the reasoning chain completely fails.\\ \cline{2-4}

  & \multirow{6}{*}{Reasoning Depth}
  & \multirow{6}{*}{\parbox{4.5cm}{\raggedright Whether the reasoning depth reflects cross-entity and hierarchical associations.}}
  & \textbf{10}: Reasoning demonstrates highly complex hierarchical relationships and deep cross-entity connections.\\ \cline{4-4}
  &&& \textbf{8--9}: Reasoning shows moderate depth; most steps are reasonable with minor ($<$10\%) missing complexity.\\ \cline{4-4}
  &&& \textbf{6--7}: Reasoning depth is insufficient; logical chains are relatively shallow, capturing only surface-level inference (10--30\% missing depth).\\ \cline{4-4}
  &&& \textbf{4--5}: Reasoning lacks depth, limited to single-layer derivations or simple restatements of facts.\\ \cline{4-4}
  &&& \textbf{1--3}: Reasoning is very superficial; most content invalid or lacks analytical depth.\\ \cline{4-4}
  &&& \textbf{0}: Reasoning chain has no depth; no higher-order inference.\\ \cline{2-4}

  & \multirow{6}{*}{Clinical Relevance}
  & \multirow{6}{*}{\parbox{4.5cm}{\raggedright Whether the reasoning contributes to diagnosis and aligns with medical context.}}
  & \textbf{10}: Reasoning fully aligns with medical context and is highly relevant and practical.\\ \cline{4-4}
  &&& \textbf{8--9}: Most reasoning is medically meaningful; only minor content is irrelevant ($<$10\%).\\ \cline{4-4}
  &&& \textbf{6--7}: Some reasoning entries are meaningful, but overall relevance is limited (10--30\% invalid content).\\ \cline{4-4}
  &&& \textbf{4--5}: Majority of content lacks medical significance; only a few entries provide support ($>$30\% clinically irrelevant).\\ \cline{4-4}
  &&& \textbf{1--3}: Reasoning is almost clinically useless; content shows deviation from medical background.\\ \cline{4-4}
  &&& \textbf{0}: Reasoning chain is entirely meaningless or invalid.\\ \cline{2-4}

  & \multirow{6}{*}{Evidence Integration}
  & \multirow{6}{*}{\parbox{4.5cm}{\raggedright Whether multiple findings and cues are integrated reasonably.}}
  & \textbf{10}: Reasoning seamlessly integrates all evidence from finding/impression chains, supporting conclusions.\\ \cline{4-4}
  &&& \textbf{8--9}: Most evidence is integrated, with minor ($<$10\%) gaps or weak concentration.\\ \cline{4-4}
  &&& \textbf{6--7}: Partial integration; some information not adopted or weakly related (10--30\%).\\ \cline{4-4}
  &&& \textbf{4--5}: Integration is poor; reasoning is limited to single evidence items ($>$30\% weak integration).\\ \cline{4-4}
  &&& \textbf{1--3}: Integration is largely insufficient; evidence shows no clear relation.\\ \cline{4-4}
  &&& \textbf{0}: Reasoning completely detached from evidence; no integration or logical coherence.\\
\hline
\end{tabular}}

\label{tab:s3_scoring}
\end{table}

\end{center}

\paragraph{Scoring Procedure.}
\textit{Input:} The ground-truth (GT) reasoning chain and the model prediction (Pred) are provided to a scoring large language model (LLM).
\textit{Processing:} Guided by a set of predefined scoring rules embedded in the evaluation prompt, the scoring LLM compares the two chains and performs a detailed analysis.
\textit{Output:} The LLM produces a structured JSON object containing the classified chains and the numerical scores for each metric, for example:\{ "scoring": \{ "s1\_finding": \{ "existence\_match": "8/10", "completeness": "7/10", "accuracy": "9/10" \}, "s2\_impression": \{...\}, "s3\_reasoning": \{...\}, "overall\_score": "xx/100" \} \}

\paragraph{Overall Score Computation.}

The final chain-of-thought evaluation score ($\text{cot\_e}$) is calculated as a weighted average of the sub-chain scores, balancing the relative importance of different reasoning levels:

\begin{equation}
CoT_e = W_{FC} \cdot \frac{1}{N}\sum^N_0(S^i_{FC}) 
               + W_{IC} \cdot \frac{1}{N}\sum^N_0(S^i_{IC}) 
               + W_{LRC} \cdot \frac{1}{N}\sum^N_0(S^i_{LRC}) ,
\end{equation}

where $S^i_{FC}$ is the score the sample $i$ in FC scoring, $S^i_{IC}$ is the score the sample $i$ in FC scoring, $S^i_{LRC}$ is the score the sample $i$ in FC scoring, and $N$ represents sample number.

The weighting coefficients satisfy:
\[
W_{FC} + W_{IC} + W_{LRC} = 1,
\]
where $W$ can be used to adjust the importance of reasoning chains during evaluation.
In this study, we set $W_{FC} =0.3$, $W_{IC} =0.3$, and $W_{LRC} =0.4$, emphasizing the importance of high-level logical reasoning (LRC) while assigning equal foundational weight to factual findings (FC) and medical impressions (IC).

\section{Cases Study}
\label{sec:appendix-cases}

\subsection{TumorChain Hallucination Correction}
In this section, we present a comparative case study to evaluate the clinical utility of TumorChain, focusing on its performance in CoT report generation. We compare the generated reports of our proposed TumorChain with those of a 2D medical model (HealthGPT) and a generalist model (Qwen2.5-VL). 

\begin{figure}
\centering
\includegraphics[width=1\textwidth]{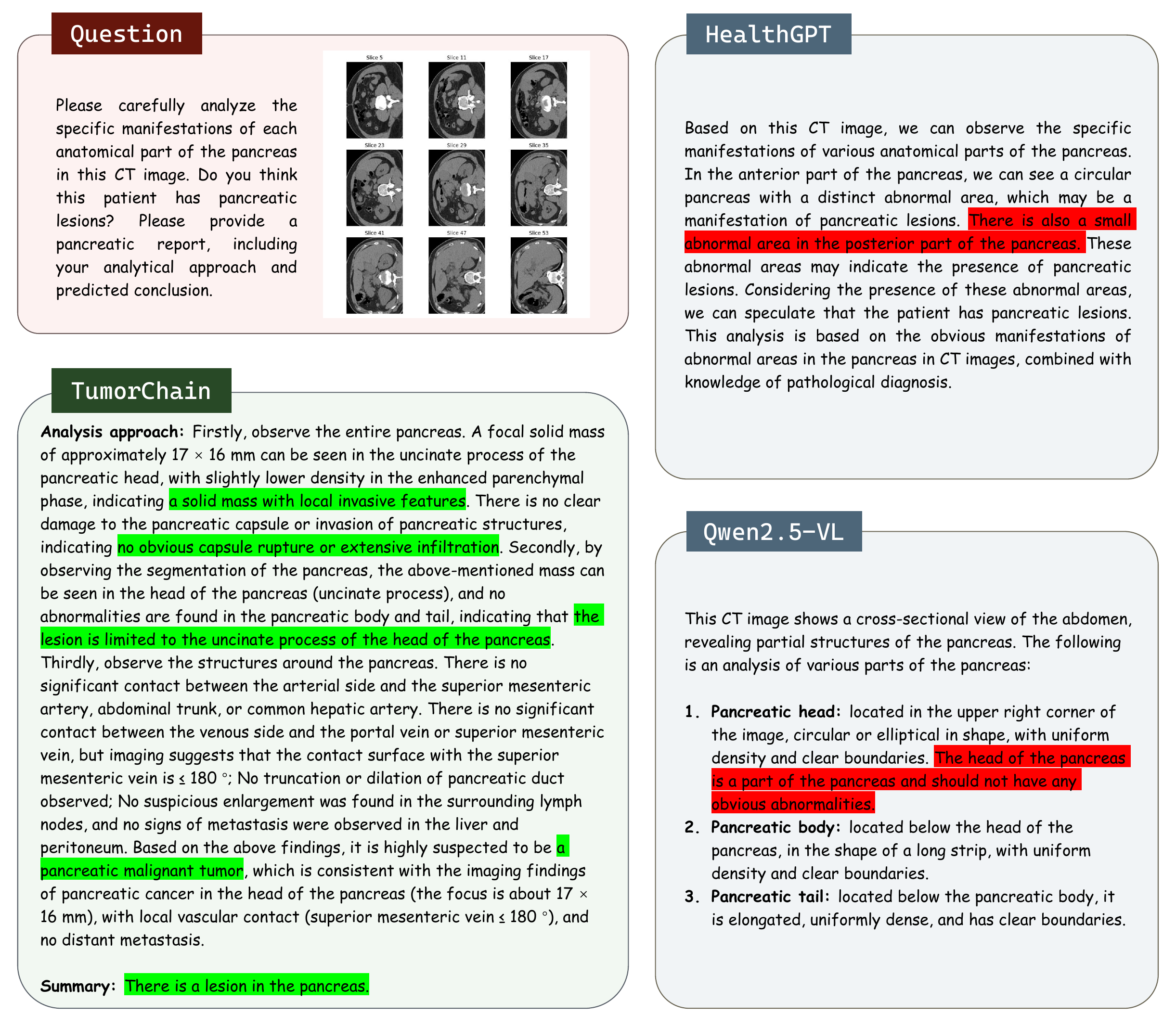}
\vspace{-2mm}
\caption{A case of CoT report generation.}
%% kp: can mention our specific technique in the caption, e.g., H$^2$LoRA, describe a bit, etc.
\vspace{-3mm}
\label{fig:case_study}
\end{figure}

Figure~\ref{fig:case_study} showcase the report outputs of the three models across a clinical scenario of a pancreatic malignant tumor, and explicitly contrast each output with the ground truth. In this example, TumorChain-7B accurately identified the location of the pancreatic tumor, abnormal density within the pancreatic parenchyma, and key features such as the capsule boundary. Although HealthGPT was extensively finetuned on general medical datasets and possesses strong image analysis capabilities, it misdiagnosed the tumor's location, incorrectly identifying it as originating from the posterior instead of the head of the pancreas. Qwen2.5-VL, meanwhile, provided an erroneous diagnosis of no abnormality in the pancreatic head.

This case demonstrates that our dataset construction process—extracting original report information by organ substructure—significantly improves the model’s ability to detect abnormalities within organ subregions. 

\subsection{Qualitative error taxonomy}

In table 1 in our paper, Tasks 1–3 correspond to fine-grained VQA pairs targeting lesion attributes. The results demonstrate that our model achieves strong performance on single-step, single-attribute tumor questions. To further analyze the sources of errors in open-ended report generation, we randomly selected 100 reports with incorrect conclusions and invited clinical experts to qualitatively review each failed case. This analysis aims to identify major error patterns and determine which lesion attributes are more prone to mistakes that could affect the final diagnosis. The main error types and representative examples are summarized as follows:

$\bullet$ \textbf{False positives due to similar CT appearances:} For example, in pancreatic cases, abnormalities such as pancreatitis or pancreatic pseudocysts are often misclassified as malignant tumors because some malignancies can also present as inflammatory changes.

$\bullet$ \textbf{Missed small lesions at organ boundaries due to anatomical overlap:} On 3D CT images, rare lesion morphologies at organ edges or corners are difficult for the model to detect. In one case, a small (approximately 2.0×1.0 cm) mixed-density lesion at the posterior edge of the pancreatic tail was missed because the overall pancreas size and surrounding anatomy appeared normal, causing the model to overlook this subtle abnormality.

$\bullet$ \textbf{Incorrect identification of the primary organ due to tumor compression:} In some cases, tumors compress adjacent organs, leading the model to misattribute the origin of the malignancy. For example, a tumor in the pancreatic tail compressing the stomach resulted in the model erroneously diagnosing a malignant lesion in the stomach.

\end{document}